\documentclass[10pt,twocolumn,letterpaper]{article}

\usepackage{cvpr}              % To produce the CAMERA-READY version
\usepackage[dvipsnames]{xcolor}

\definecolor{cvprblue}{rgb}{0.21,0.49,0.74}
\usepackage[pagebackref,breaklinks,colorlinks,citecolor=cvprblue]{hyperref}
\usepackage{graphicx}
\usepackage{amsmath}
\usepackage{amssymb}
\usepackage{booktabs}
\usepackage{diagbox}
\usepackage{microtype}      % microtypography
\usepackage{graphicx}
\usepackage{float}
\usepackage{bbding}
\usepackage{enumitem}
\def\confName{CVPR}
\def\confYear{2025}

\usepackage{epsfig}
\usepackage{multicol}
\usepackage{multirow}

\usepackage[capitalize]{cleveref}
\crefname{section}{Sec.}{Secs.}
\Crefname{section}{Section}{Sections}
\Crefname{table}{Table}{Tables}
\crefname{table}{Tab.}{Tabs.}

\def\authorBlock{
    Guanglu Dong\textsuperscript{1} \qquad
    Tianheng Zheng\textsuperscript{1} \qquad
    Yuanzhouhan Cao\textsuperscript{2}   \qquad
    Linbo Qing\textsuperscript{1} \qquad
    Chao Ren\textsuperscript{1}\thanks{Corresponding author} %\footnotemark[1]
     \\

        \textsuperscript{1}Sichuan University, Chengdu, China \qquad
    \textsuperscript{2}Beijing Jiaotong University, Beijing, China\\

	{\tt\small \{dongguanglu, zhengtianheng\}@stu.scu.edu.cn, }
    {\tt\small yzhcao@bjtu.edu.cn}\\
	{\tt\small \{qing\_lb, chaoren\}@scu.edu.cn}
    
}

\title{Channel Consistency Prior and Self-Reconstruction Strategy Based Unsupervised Image Deraining}
\author{\authorBlock}
\begin{document}
\maketitle
\renewcommand{\thefootnote}{\fnsymbol{footnote}}

\begin{abstract}
Recently, deep image deraining models based on paired datasets have made a series of remarkable progress. However, they cannot be well applied in real-world applications due to the difficulty of obtaining real paired datasets and the poor generalization performance. In this paper, we propose a novel \textbf{C}hannel Consistency Prior and \textbf{S}elf-Reconstruction Strategy Based \textbf{U}nsupervised Image \textbf{D}eraining framework, \textbf{CSUD}, to tackle the aforementioned challenges. During training with unpaired data, CSUD is capable of generating high-quality pseudo clean and rainy image pairs which are used to enhance the performance of deraining network. Specifically, to preserve more image background details while transferring rain streaks from rainy images to the unpaired clean images, we propose a novel Channel Consistency Loss (CCLoss) by introducing the Channel Consistency Prior (CCP) of rain streaks into training process, thereby ensuring that the generated pseudo rainy images closely resemble the real ones. Furthermore, we propose a novel Self-Reconstruction (SR) strategy to alleviate the redundant information transfer problem of the generator, further improving the deraining performance and the generalization capability of our method. Extensive experiments on multiple synthetic and real-world datasets demonstrate that the deraining performance of CSUD surpasses other state-of-the-art unsupervised methods and CSUD exhibits superior generalization capability. Code is available at \href{https://github.com/GuangluDong0728/CSUD-Unsupervised-Deraining-CVPR2025}{https://github.com/GuangluDong0728/CSUD}.
\end{abstract}

\section{Introduction}
\label{sec:intro}

\label{Introduction}
Image deraining is an important task in low-level computer vision \cite{zunei_SR, zunei_denoising, DerainNet, darkchannel, restormer, PromptIR}. It aims to reconstruct a clean image from its rainy version, which has a profound impact on subsequent high-level tasks such as object detection, recognition, and semantic segmentation \cite{ref1,Detection,ref2,zunei_Segmentation}. To mitigate the negative impact of rain, numerous deep learning based supervised image deraining methods \cite{DerainNet, DDN, DIDMDN, ref3, RESCAN, PReNet, SPDNet, SPANet, MSPFN, SEIDNet} have been proposed and they achieve remarkable progress on synthetic datasets. However, the performance of these methods is largely constrained by the difficulty of capturing a large number of high-quality paired datasets when applied to real-world scenarios. Additionally, attempting to directly apply supervised models trained on synthetic datasets to real-world scenarios is also difficult because of the huge domain gap between synthetic and real-world rain \cite{Closing} and the overfitting issue of supervised methods \cite{generalization}.

\begin{figure}[t]
  \centering
  \setlength{\abovecaptionskip}{0.1cm}
  \includegraphics[width=0.48\textwidth]{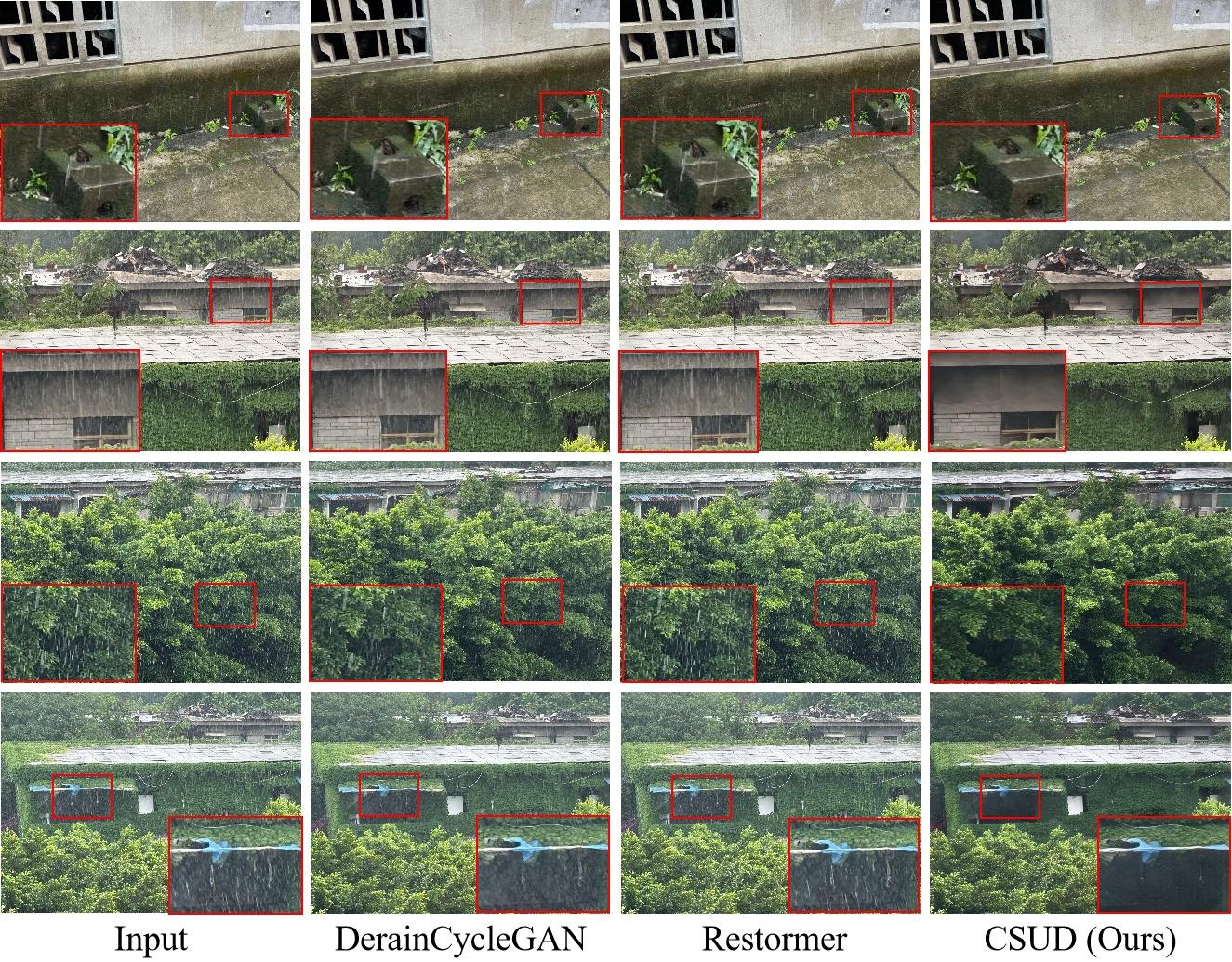}
  %\fbox{\rule[-.5cm]{0cm}{4cm} \rule[-.5cm]{4cm}{0cm}}
  \caption{Deraining results on the real rainy images captured by ourselves in real-world scenarios. Compared with the supervised method Restormer \cite{restormer} and the unsupervised method DerainCycleGAN \cite{DerainCycleGAN}, \textbf{our CSUD exhibits extremely strong generalization capability in the real world and achieves the best visual results.}}
  % More visual results are presented in the Appendix.
  \label{fig.realshot}
\end{figure}
% which both fail to clearly remove the rain streaks

Compared to supervised methods, unsupervised deep learning methods \cite{ICIP, DerainCycleGAN, DCDGAN, NLCL, QTP} for image deraining tasks generally achieve end-to-end image deraining without paired clean-rainy images during training. This allows unsupervised methods to effectively handle complex real-world scenarios, especially when data collection costs are high. However, due to the high difficulty and challenges of unsupervised deraining methods in training process and the lack of prior knowledge \cite{DerainCycleGAN}, there is still very little relevant research in this field. Additionally, the deraining performance of existing unsupervised methods always lags far behind that of supervised methods, and they face the same generalization issues as supervised methods. 

As shown in \cref{fig.realshot}, even the SOTA supervised and unsupervised deraining models \cite{restormer, DerainCycleGAN} encounter substantial generalization issues and struggle to achieve satisfactory results when dealing with rainy images captured in real world, and sometimes they even fail to remove any rain streaks due to the more complex rain distributions in real world. Based on the above analysis, how to design an effective unsupervised deraining framework to avoid the need for real paired data and how to improve the generalization performance of deraining methods have become key research issues.

% All methods are trained on synthetic datasets.

% our method addresses the challenge of training unsupervised frameworks with unpaired data and obtains extraordinary synthetic-to-real generalization performance. Specifically,

In this paper, we propose a novel channel consistency prior and self-reconstruction strategy based unsupervised image deraining framework which only needs unpaired data to train the network and can be easily generalized to real-world scenarios. We first conduct detailed analysis of the channel consistency prior (CCP) of rain streaks, and then propose a novel channel consistency loss to constrain the generator to synthesize high-quality pseudo rainy images. In addition, we utilize the proposed self-reconstruction (SR) strategy to further improve the performance of the generator and enhance the generalization capability of the derainer. As shown in Figure \ref{fig.realshot}, our CSUD achieves the most promising performance on the real-world captured images compared to both supervised and unsupervised deraining methods. The main contributions of this work are summarized below:
\begin{itemize}[leftmargin=*]
\item
We propose a novel channel consistency prior and self-reconstruction strategy based unsupervised image deraining framework which is able to generate high-quality pseudo paired data for training. This framework can effectively address the issues of insufficient paired data in real world and significantly improve the generalization capability of the network.
\item
We propose a novel channel consistency loss (CCLoss) according to the channel consistency prior (CCP) of rain streaks to enhance the performance of the generator, preserving more background details while generating pseudo rainy images. To the best of our knowledge, we are the first to incorporate CCP into the training process of unsupervised deraining framework. 
\item
We propose a self-reconstruction (SR) strategy to solve the redundant information transfer problem of the generator. SR strategy can facilitate the accuracy of rain streaks transfer in the generator and enhance the deraining performance and the generalization capability of the derainer without altering the existing network structure.
\item
Extensive experiments on numerous benchmark datasets demonstrate that our proposed unsupervised framework can effectively tackle the deraining tasks on both synthetic and real-world rainy images with extremely superior generalization capability in real-world scenarios.
\end{itemize}

\section{Related Works}
\label{Related Works}

\begin{figure*}[htb]
  \centering
  \setlength{\abovecaptionskip}{0.1cm}
  \includegraphics[width=1\textwidth]{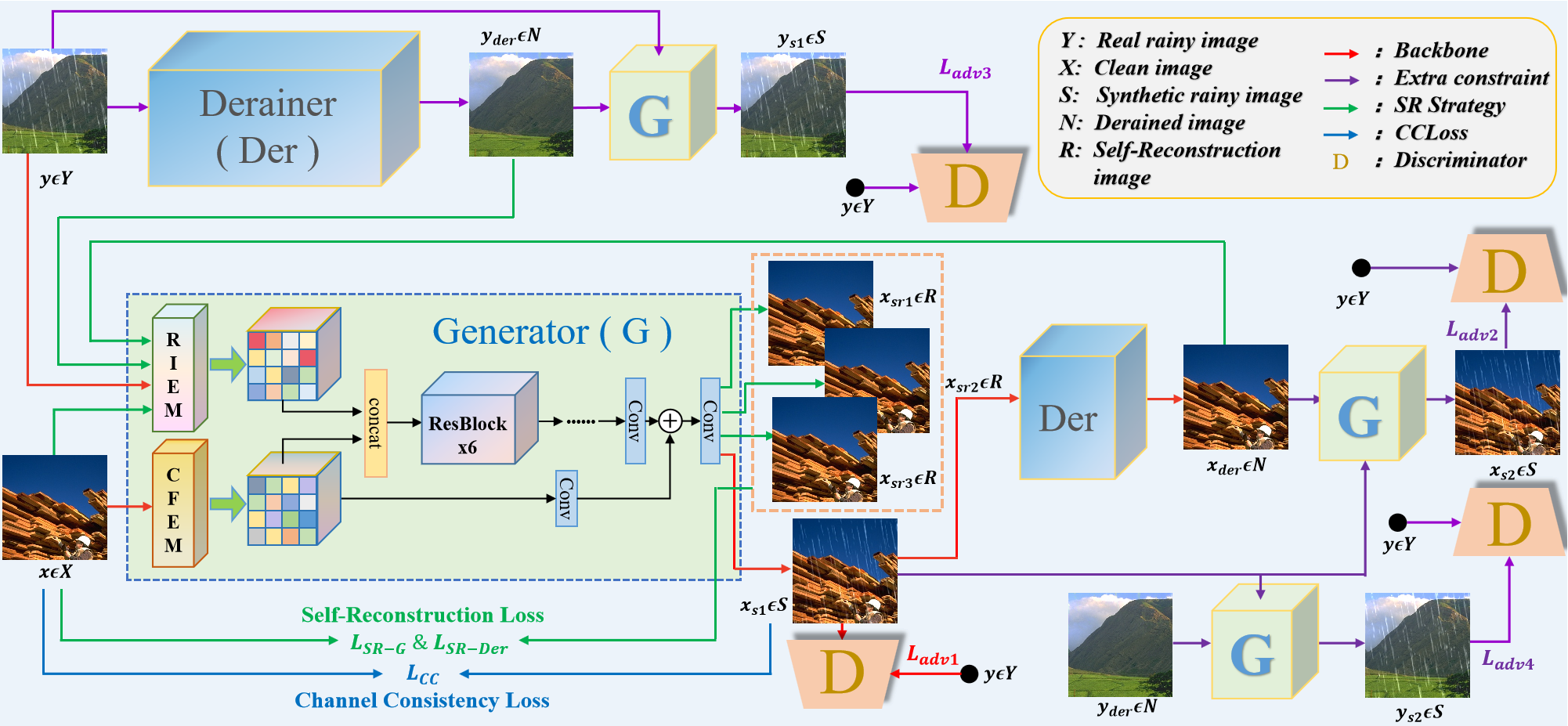}
  %\fbox{\rule[-.5cm]{0cm}{4cm} \rule[-.5cm]{4cm}{0cm}}
  \caption{Framework of the proposed CSUD. The red arrows represent the baseline, purple arrows represent the 3 extra constraints, green arrows represent the self-reconstruction (SR) strategy, and blue arrows represent the CCLoss. The 4 input arrows of RIEM represent 4 separate runs of the generator, each time guided by a different image. Both RIEM and CFEM receive only 1 input image at a time.}
  \label{fig.Framework}
\end{figure*}

\subsection{Deep Learning Based Single Image Deraining}
\textbf{Supervised Single Image Deraining.} 
Recently, with the development of deep learning, numerous excellent single image deraining \cite{DDN, DerainNet, DIDMDN, DRSformer, MSPFN, NeRD, SEIDNet, SPDNet, SPANet} models have emerged. Fu et al. \cite{DerainNet} first employ the CNN-based DerainNet to extract and remove the rain layer. Lin et al. \cite{SEIDNet} propose a generative network equipped with the pixel-wise status estimation and information decoupling for deraining. Lin et al. \cite{LowLightDeraining} propose a unified network for low-light-rainy image restoration to jointly handle low-light enhancement and deraining tasks. In addition to the aforementioned works specifically targeting deraining, some networks designed for image restoration \cite{MPRNet, restormer, PromptRestorer, PromptIR, AirNet, HINet, Uformer, allinone} have also shown excellent performance in deraining tasks. However, the above fully supervised methods require a large number of high-quality paired training samples, which is almost impossible in real-world scenarios. Simultaneously, due to the overfitting issue, their generalization capability is limited. 

% Spatial attentive network (SPANet) \cite{SPANet} extracts the spatial contextual information based on the recurrent network and obtains the spatial details from local to global. 

% \noindent
\textbf{Unsupervised Single Image Deraining.} To address the lack of paired datasets in the real world, some outstanding unsupervised deraining methods \cite{DCDGAN, DerainCycleGAN, ICIP, NLCL, QTP} have been proposed. Wei et al. \cite{DerainCycleGAN} propose an unsupervised method for rain removal and generation by utilizing the popular CycleGAN \cite{CycleGAN} framework. Then, through introducing a dual contrastive learning manner to the CycleGAN framework, Chen et al. \cite{DCDGAN} develop an unpaired adversarial framework which effectively explores mutual properties of the unpaired rainy-clean samples. Ye et al. \cite{NLCL} introduce a novel non-local contrastive learning based unsupervised image deraining method to better distinguish rain streaks from clean images. However, the deraining performance of these unsupervised methods is far inferior to that of supervised methods, especially on real-world scenarios. Compared with previous methods, our proposed CSUD can greatly enhance the unsupervised deraining performance and achieve competitive results compared with some supervised methods.

% Xu et al. \cite{QTP} propose a hybrid quality-task-perception loss to learn a better unsupervised image restoration model, which has shown superiority in deraining task.

\subsection{Generalization Problem in Image Deraining}
Deep image deraining models often fail to obtain satisfactory performance or even cannot remove any rain streaks when they are applied to real-world scenarios. Generalization problem of deraining models is an urgent issue to be solved. Gu et al. \cite{generalization} conduct thorough analysis of the generalization problem in image deraining and propose to achieve better generalization by simplifying the complexity of training images. Some semi-supervised learning frameworks \cite{SIRR, S2R} are proposed to analyze the residual difference between domains and enhance the deraining capability in real world. In addition, a series of unsupervised deraining methods \cite{DerainCycleGAN,DCDGAN,NLCL} have also demonstrated a certain degree of generalization capability in real-world scenarios. However, they have not conducted detailed analysis on how to enhance the generalization capability of unsupervised methods. Inspired by \cite{generalization}, we propose a self-reconstruction strategy to make both generator and derainer pay more attention to the reconstruction of image background, which significantly improves the generalization performance of our method.

\section{Methodology}
\label{Methodology}
% We provide a detailed explanation of the proposed unsupervised image deraining framework in this section. Our method aims to address the issue of insufficient paired data in real-world deraining applications, and significantly improve the generalization performance of the network. 
% Additionally, we proposed a channel consistency loss (CCLoss) and a self-reconstruction (SR) strategy to further improve the deraining performance and the generalization capability of CSUD. CCLoss and SR strategy will be explained detailedly in Sec \ref{sec.rgbloss} and Sec \ref{sec.sr}.

\begin{figure*}[htb]
  \centering
  \setlength{\abovecaptionskip}{0.1cm}
  \includegraphics[width=1\textwidth]{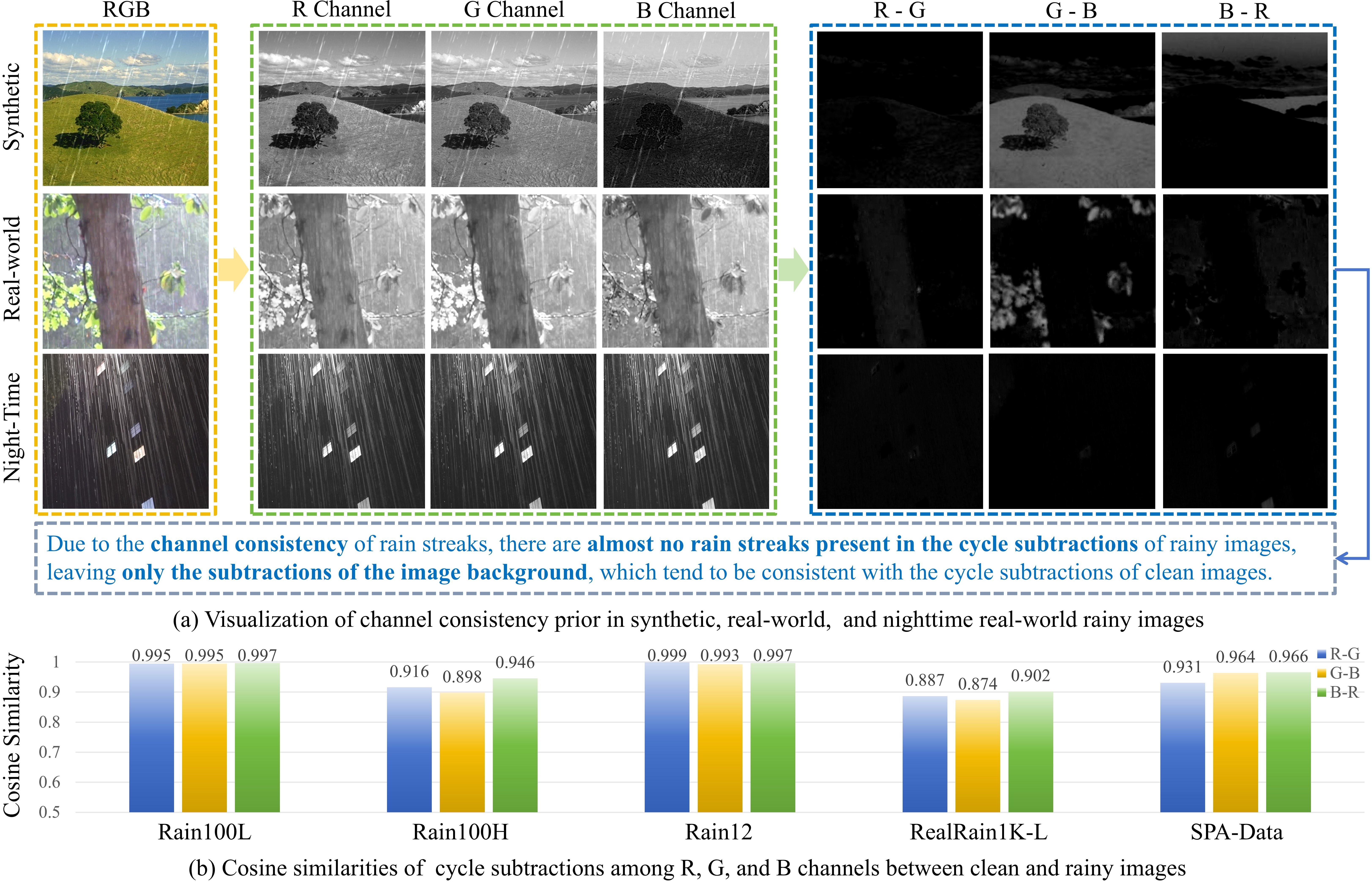}
  %\fbox{\rule[-.5cm]{0cm}{4cm} \rule[-.5cm]{4cm}{0cm}}
  \caption{(a) Visualization of channel consistency prior in synthetic and real-world rainy images. The first column presents the rainy RGB images; the second, third and fourth columns present their R, G, B channels, respectively; the fifth, sixth, and seventh columns present the cycle subtractions of R, G, and B channels, respectively. (b) Cosine similarities of the cycle subtractions of R, G, and B channels between clean and rainy images in the 5 synthetic and real-world datasets. \textbf{More visualizations and analysis are presented in the Appendix}.}
  \label{fig.rgbloss}
\end{figure*}

\subsection{Overall Framework of Proposed Method}
To address the issue of insufficient paired data and the generalization problem in deraining applications, we propose a novel channel consistency prior and self-reconstruction strategy based unsupervised image deraining framework, CSUD. As illustrated in \cref{fig.Framework}, our CSUD consists of a derainer $Der$, a rainy image generator $G$, and a Patch-GAN \cite{PatchGAN} discriminator $D$. For $G$, it consists of a clean feature extraction module (CFEM), a rain information extraction module (RIEM) and 6 residual blocks. \textbf{Detailed network structures are detailed in the Appendix.}
% \textbf{The detailed structures are presented in the Appendixs.} 

% CFEM and RIEM respectively denote the clean feature extraction module and rain information extraction module.

In generator $G$, we effectively utilize unpaired clean and rainy images to synthesize pseudo rainy images. As shown by the red arrows in \cref{fig.Framework}, we design a GAN-based baseline that transitions from “$x \epsilon X$ $\rightarrow$ $x_{s1} \epsilon S$ $\rightarrow$ $x_{der} \epsilon N$”, where $X$, $S$ and $N$ denote clean images, synthetic rainy images and derained images respectively. $G$ transforms clean image $x$ to synthetic rainy image $x_{s1}$ according to the guide of real rainy image $y$. In order to better extract rain information and transfer it to the clean images, $y$ and $x$ are initially processed by the RIEM and the CFEM respectively to obtain rainy features and clean features which are concatenated for subsequent process. For the real rainy image $y$ and the synthesized rainy image $x_{s1}$, the corresponding derained clean images $y_{der}$ and $x_{der}$ are obtained through $Der$.

The discriminator $D$ is trained to distinguish between synthetic rainy images $x_{s1} \epsilon S$ and real rainy images $y \epsilon Y$, and the adversarial loss ${L}_{adv1}$ of the baseline is constrained between them. Furthermore, motivated by existing methods \cite{CycleGAN, LinDenoising}, as indicated by the purple arrows in \cref{fig.Framework}, we add another three processes, $G(Der(x_{s1}), x_{s1}) \rightarrow x_{s2}$, $G(Der(y), y) \rightarrow y_{s1}$, and $G(Der(y), x_{s1}) \rightarrow y_{s2}$, to the framework to enhance training stability. Then, additional three adversarial losses are introduced to CSUD:
\[ 
\begin{aligned}
{L}_{GAN} = {L}_{adv1} + {L}_{adv2} + {L}_{adv3} + {L}_{adv4} .
\end{aligned}
\tag{1}
\] 
% Additionally, we proposed a channel consistency loss (CCLoss) and a self-reconstruction (SR) strategy to further improve the deraining performance and the generalization capability of CSUD. CCLoss and SR strategy will be explained detailedly in Sec \ref{sec.rgbloss} and Sec \ref{sec.sr}.

% \textbf{Generation of Pseudo Rainy Images and Reconstruction of Derainer.} 

% The RIEM is based on a U-Net architecture, comprising a downsampling layer followed by an upsampling layer, while the CFEM simply utilize a convolutional layer. The generator $G$ learns the rain characteristics of rainy images to guide the synthesis of clean images towards realistic rainy ones. 

\subsection{Channel Consistency Loss for Generator} 
\label{sec.rgbloss}

% \noindent
\textbf{Channel Consistency Prior (CCP) of Rain Streaks.} 
In RGB rainy images, most rain streaks tend to be consistent across the R, G, and B channels, meaning the pixel values of most rain streaks in the three channels are highly similar. As shown in the second, third, and fourth columns of \cref{fig.rgbloss} (a), we separate both synthetic and real-world rainy images into R, G, and B channels and visualized them. It can be seen clearly that while there are significant differences in the background parts among the three channels, the rain streaks tend to be consistent. We propose to term this phenomenon as "Channel Consistency Prior (CCP)" of rainy streaks. 

% \begin{figure*}[htb]
%   \centering
%   \setlength{\abovecaptionskip}{0.cm}
%   \includegraphics[width=1\textwidth]{fig/SRStrategy.png}
%   %\fbox{\rule[-.5cm]{0cm}{4cm} \rule[-.5cm]{4cm}{0cm}}
%   \caption{(a) Visualization of redundant information transfer problem. The first and second columns present the input rainy and clean images of the generator, respectively; the third and fourth columns present pseudo rainy images generated by the generator without/with SR strategy, respectively. (b) The implementation principles of the proposed SR strategy.}
%   \label{fig.SRStrategy}
% \end{figure*}

Theoretically, according to CCP, the cycle subtractions $((r)_{R} - (r)_{G})$, $((r)_{G} - (r)_{B})$, $((r)_{B} - (r)_{R})$ of R, G, and B channels in rainy images will almost contain no rain streaks and should be close to $((n)_{R} - (n)_{G})$, $((n)_{G} - (n)_{B})$, $((n)_{B} - (n)_{R})$ in clean images, where $r$ and $n$ represent the rainy and clean image, $(\cdot)_{R}$, $(\cdot)_{G}$, and $(\cdot)_{B}$ denote R, G, and B channels respectively. In the fifth, sixth, and seventh columns of \cref{fig.rgbloss} (a), it can be observed that the cycle subtractions of R, G, and B channels nearly don't contain any rain streaks, qualitatively confirming the correctness of CCP. Furthermore, as shown in \cref{fig.rgbloss} (b), we quantitatively present the cosine similarities of the cycle subtractions of R, G, and B channels between clean and rainy images in both synthetic and real-world benchmark datasets \cite{Rain100L,realrain1k,SPANet,Rain12}. It is evident that the cosine similarities are exceptionally high, which can further demonstrate our hypothesis.

% \noindent
\textbf{Channel Consistency Loss.} Based on the above analysis of CCP, the cycle subtractions of R, G, and B channels between input clean image and output pseudo rainy image of the generator $G$ should tend to be consistent. To this end, we propose a novel Channel Consistency Loss (CCLoss) to preserve more color and texture details of the background while generating pseudo rainy images:
\[ 
\begin{aligned}
L_{CC} = &\left | \left | ((x_{s1})_{R} - (x_{s1})_{G}) ,  (x_{R} - x_{G})\right | \right |_{1} \\ &+ \left | \left | ((x_{s1})_{G} - (x_{s1})_{B}) ,  (x_{G} - x_{B})\right | \right |_{1} \\ &+ \left | \left | ((x_{s1})_{B} - (x_{s1})_{R}) ,  (x_{B} - x_{R})\right | \right |_{1} ,
\end{aligned}
\tag{2}
\]
where $(x_{s1})_{R}$, $(x_{s1})_{G}$, $(x_{s1})_{B}$, $x_{R}$, $x_{G}$, $x_{B}$ denotes R, G, B color channels of $x_{s1}$ and $x$, $ \left | \left | \cdot \right | \right |_{1}$ represents the $L_1$ norm and $L_{CC}$ represents the proposed CCLoss, respectively. By introducing CCP into our unsupervised deraining framework, $G$ will generate more accurate synthesized rainy images, which will further improve the performance of the derainer.

\subsection{Self-Reconstruction Strategy}
\label{sec.sr}
\begin{figure}[t]
  \centering
  \setlength{\abovecaptionskip}{0.1cm}
  \includegraphics[width=0.48\textwidth]{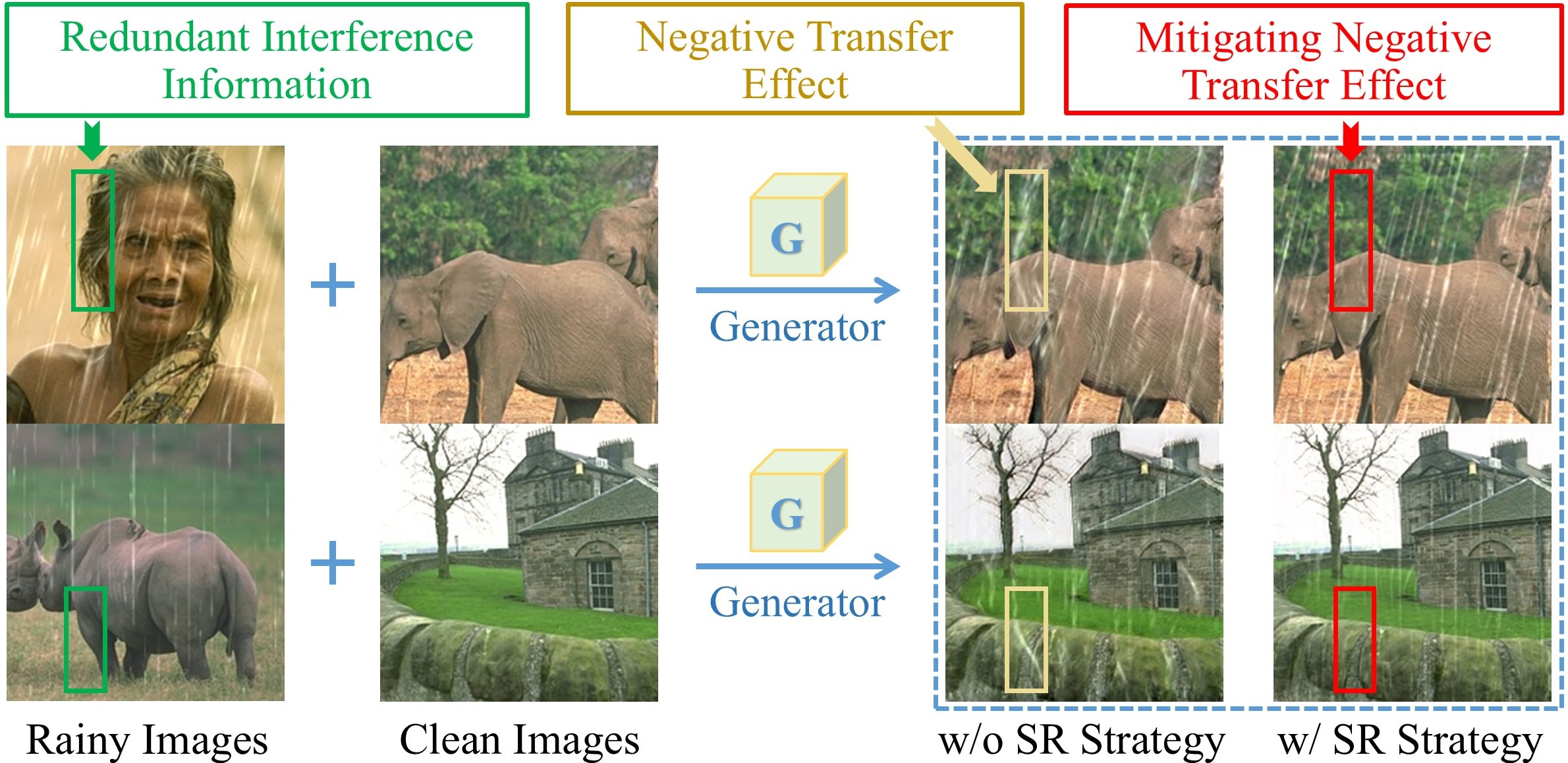}
  %\fbox{\rule[-.5cm]{0cm}{4cm} \rule[-.5cm]{4cm}{0cm}}
  \caption{Visualization of redundant information transfer problem.}
  \label{fig.SRStrategy1}
\end{figure}

% \noindent
\textbf{Redundant Information Transfer Problem of the Generator.} Although adversarial losses can effectively constrain the training process of generator $G$, the data synthesis process of $G$ is highly susceptible to interference from the redundant information in rainy images, especially those edge textures that are very similar to rain streaks. As shown in \cref{fig.SRStrategy1}, the edge textures of some background objects in rainy images, such as the outline of the woman's hair, the leg of rhino, etc., are mistakenly transferred to the clean images by $G$. As a result, $G$ will generate inaccurate pseudo rainy images, interfering with the learning process of $Der$. 

% \noindent
\textbf{Self-Reconstruction Strategy.} To solve the redundant information transfer problem, we propose a novel Self-Reconstruction (SR) strategy, which enables $G$ to extract more accurate rain streak information and synthesizes more realistic pseudo rainy images. The core idea of SR strategy is to encourage $G$ to better distinguish rain streaks from background in the rainy images, ensuring that only rain streak information is extracted for subsequent generation of pseudo rainy images, while redundant information of background is disregarded. As shown by the green arrows in \cref{fig.Framework}, when the input clean image $x$ itself, the derained image $y_{der}$ are input with clean image $x$ into $G$, $G$ learns to reconstruct the input clean image $x$ itself without any additional redundant information from $x$ and $y_{der}$, which means that the self-reconstructed images $x_{s1}$ and $x_{s2}$ should be close to $x$. Thus, we can utilize the following self-reconstruction loss (SRLoss) as additional training constraint of $G$:
\[ 
L_{SR-G} = \left | \left | G(x, x), x \right | \right|_{1} + \left | \left | G(x, y_{der}), x \right | \right|_{1} ,
\tag{3}
\]
where $L_{SR-G}$ represents the SRLoss for $G$. SR strategy allows $G$ to pay more attention to reconstructing the background of clean images while learning the transfer of rain streaks, significantly improving the accuracy of generating pseudo rainy images. Under the constraint of $L_{SR-G}$, $G$ will obtain the ability to generate clean images themselves guided by images without rain streaks. Therefore, the cleaner the derained images $y_{der}$ and $x_{der}$ restored by $Der$ are, the closer the self-reconstructed images $x_{s2}$ and $x_{s3}$ are to the original clean ones $x$. Then, the following SRLoss can be utilized to further improve the performance of $Der$:
\[ 
L_{SR-Der} = \left | \left | G(x, x_{der}), x \right | \right|_{1} + \left | \left | G(x, y_{der}), x \right | \right|_{1} ,
\tag{4}
\]
where $L_{SR-Der}$ represents the SRLoss for $Der$. Under the constraint of $L_{SR-Der}$, $Der$ will also focus more on reconstructing the background content while learning to derain, which greatly enhances the generalization capability of $Der$.

\begin{figure}[t]
  \centering
  \setlength{\abovecaptionskip}{0.2cm}
  \includegraphics[width=0.48\textwidth]{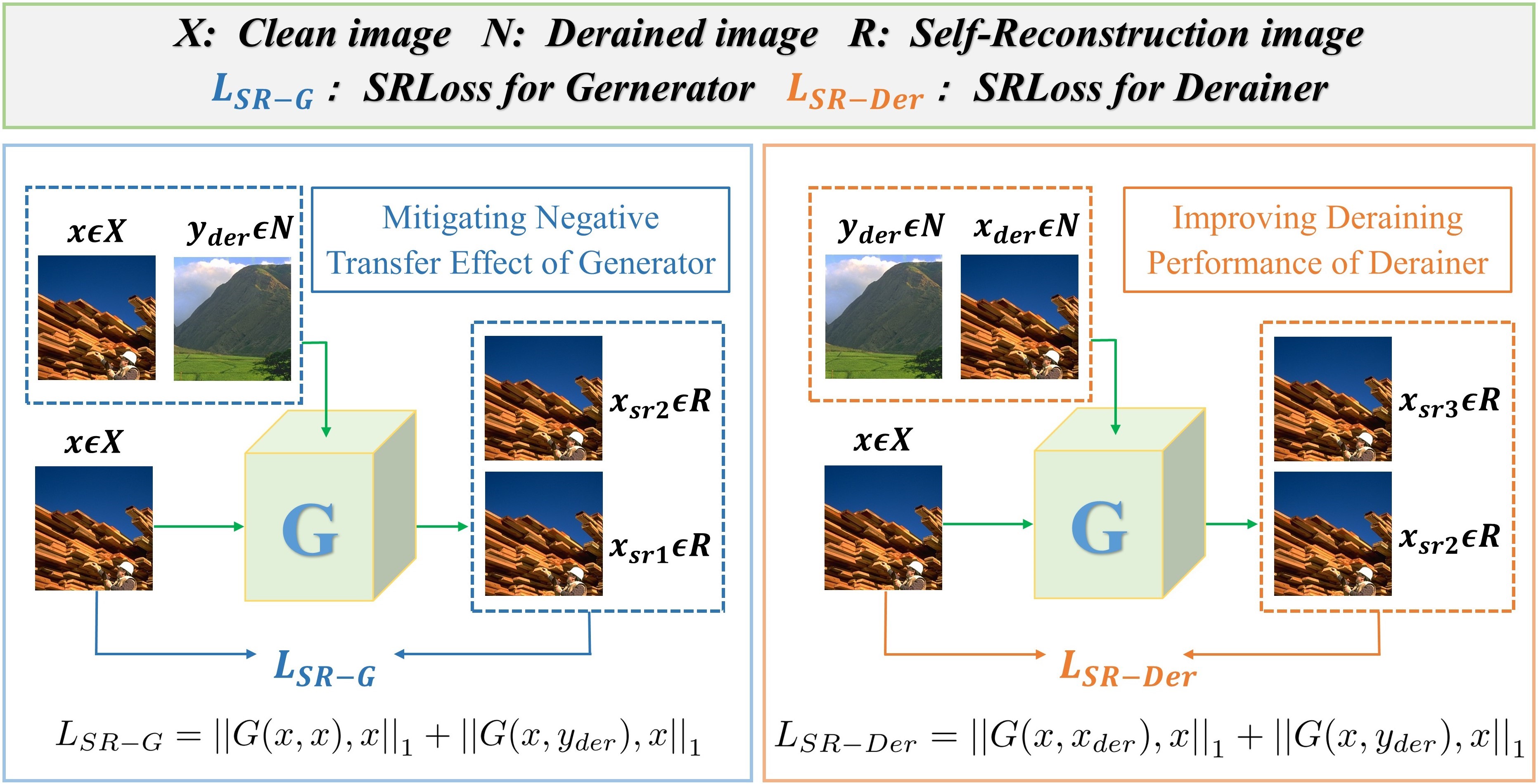}
  %\fbox{\rule[-.5cm]{0cm}{4cm} \rule[-.5cm]{4cm}{0cm}}
  \caption{The implementation principles of SR strategy.}
  \label{fig.SRStrategy2}
\end{figure}

As shown in the third and fourth columns of Figure \ref{fig.SRStrategy1}, SR strategy effectively alleviates the redundant information transfer problem. Through SR strategy, $G$ can achieve robust rain streak transfer with more background detail fidelity, greatly enhancing the deraining performance of $Der$.

% \begin{figure}[htb]
%   \centering
%   \label{fig.SRStrategy}
%     \begin{minipage}{0.58\textwidth} 
%       % \flushleft
%       % \centering
%       % \setlength{\abovecaptionskip}{0.cm}
%       \includegraphics[width=1\textwidth]{fig/SRStrategy.png}
%     \end{minipage}
%   %\fbox{\rule[-.5cm]{0cm}{4cm} \rule[-.5cm]{4cm}{0cm}}
%     \begin{minipage}{0.38\textwidth} 
%       % \centering
%       \caption{Visualization of redundant information transfer problem. The first and second columns represent the input rainy and clean images of the generator, respectively; the third and fourth columns represent pseudo rainy images generated by the generator without/with the self-reconstruction strategy, respectively.}
%     \end{minipage}
% \end{figure}

\subsection{Loss Function}

% Due to the high difficulty and instability of training a GAN framework, only using loss function such as L1 loss to restrict the optimization process of $Der$ will likely lead to training collapse. Therefore, to alleviate this issue, in addition to adopting the L1 loss for $Der$, we also incorporate SSIM loss and perceptual loss\cite{Perceptual} to enhance the reconstruction performance of $Der$: 
Due to the high difficulty and instability of training a GAN-based unsupervised framework, in addition to adopting the L1 loss for $Der$, we also incorporate SSIM loss and perceptual loss\cite{Perceptual} to enhance the performance of $Der$: 
\[ 
\begin{aligned}
L_{Der} &= \left | \left | x, x_{der} \right | \right |_{1} + \lambda_{1} L_{SSIM}(x, x_{der}) \\ &+ \lambda_{2}\left | \left | \phi_{i}(x),  \phi_{i}(x_{der}) \right | \right |_{1} + \lambda_{3}L_{SR-Der} , \label{eq12}
\end{aligned}
\tag{5}
\]
where $ \left | \left | \cdot \right | \right |_{1}$ represents the $L_1$ norm, and $L_{SSIM}(\cdot)$ represents the SSIM loss, $\phi_{i}(\cdot)$ is the layers of pre-trained VGG-19 \cite{VGG}, $\lambda_{1}$, $\lambda_{2}$,  $\lambda_{3}$ represent the hyperparameters for SSIM loss, perceptual loss and SRLoss for derainer respectively.

Finally, we obtain the total loss to train our framework: 
\[ 
\begin{aligned}
L_{total} = \mathop{min}\limits_{G}\mathop{max}\limits_{D}L_{GAN} + L_{Der} +  \alpha_{1} L_{CC} + \alpha_{2} L_{SR-G} , \label{eq13}
\end{aligned}
\tag{6}
\]
where $\alpha_{1}$, $\alpha_{2}$ represent the hyperparameters for CCLoss and SRLoss for generator respectively.

\begin{table*}[htb]
    \normalsize
      \caption{Quantitative deraining performance comparisons of different methods. Red and blue colors indicate the 1st and 2nd ranks among the unsupervised methods, respectively. \textbf{S and U indicate supervised method and unsupervised method, respectively.}}
      % \resizebox{\textwidth}{!}{
      \label{table.Unsupervisedderainingresults}
      \centering
      \scalebox{0.75}{
      \begin{tabular}{c|cccccccccccc}
        \toprule
        \multicolumn{1}{c|}{Datasets} & \multicolumn{2}{c}{Rain100L \cite{Rain100L}} & \multicolumn{2}{c}{Rain100H \cite{Rain100L}} & \multicolumn{2}{c}{Rain12 \cite{Rain12}} & \multicolumn{2}{c}{Rain800 \cite{Rain800}} & \multicolumn{2}{c}{RealRain1K-L \cite{realrain1k}} & \multicolumn{2}{c}{RealRain1K-H \cite{realrain1k}}                      \\
        % \cmidrule(r){1-2}
        \midrule
        \multicolumn{1}{c|}{Metrics} & PSNR & SSIM & PSNR & SSIM & PSNR & SSIM & PSNR & SSIM  & PSNR & SSIM & PSNR & SSIM               \\
        \midrule
                                DerainNet \cite{DerainNet} (S)&27.03& 0.884&  14.92 &0.592& /&/&22.77 &0.810&27.09 &0.925&22.88 &0.889\\
                                DDN \cite{DDN} (S)  & 32.38&0.926&24.64&0.849&34.04&0.933&21.16&0.732&31.18&0.917&29.17&0.878     \\
                                  % DID-MDN \cite{DIDMDN} (S) &35.40&0.961&25.61&0.854&/&/&21.89&0.795 &/&/&/&/      \\
      SPA-Net \cite{SPANet} (S)       &31.95&0.919&26.07&0.857&/&/&24.37&0.861&30.43&0.947&25.76&0.910 \\           
                           MPRNet \cite{MPRNet} (S) &34.95&0.959&28.53&0.872&36.84&0.964&28.68&0.875&36.29&0.972&34.74&0.964\\    
                                 NAFNet \cite{NAFNet} (S)  &37.00&0.978&29.66&0.900&34.81&0.943&28.72&0.886&38.80&0.986&36.11&0.976   \\
                                  Restormer \cite{restormer} (S)    &37.57&0.974&29.46&0.889&/&/&30.33&0.905 &40.90&0.985&39.57&0.981  \\
                                  PromptIR \cite{PromptIR} (S)    &38.34&0.983&28.69&0.877&35.09&0.945&29.07&0.887 &36.99&0.973&33.61&0.953  \\
                                NeRD-Rain-S \cite{NeRD} (S)    &42.00&0.990& 32.86 & 0.932 &35.39&0.942& 28.82& 0.878 & 38.64&0.979 & 36.69&0.970  \\
        % \midrule
        %                           SIRR\cite{SIRR}(M) & 34.47  &0.969&26.55&0.846&/&/&24.36&0.859&/&/&/&/  \\
        % % SS-TL(M)  &  32.37&0.926&/&/&34.02&0.935 &/&/&/&/&/&/   \\
        %                       Syn2Real\cite{S2R}(M) &34.39&0.965&25.76&0.837&/&/&23.74&0.799&/&/&/&/   \\
        \midrule
                                  CycleGAN \cite{CycleGAN} (U) & 24.61 & 0.834 &20.59&0.704&21.56&0.845&23.95& \textcolor{blue}{0.819} &20.19&0.820&17.53&0.762 \\                            
               DerainCycleGAN \cite{DerainCycleGAN} (U)   &  31.49 &  0.936 &  21.13 & 0.710 &\textcolor{blue}{33.52}&\textcolor{blue}{0.940}&\textcolor{blue}{24.32}&\textcolor{red}{0.842}&28.16&0.901&24.78&0.844  \\   
               NLCL \cite{NLCL} (U)    & 20.50 & 0.719 &22.31&0.728&22.68&0.735&18.94&0.670&23.06&0.832&  20.64 &0.719\\
        DCD-GAN \cite{DCDGAN} (U) &  \textcolor{blue}{31.82}&\textcolor{blue}{0.941}&\textcolor{blue}{22.47}&\textcolor{blue}{0.753}&31.56&0.924&\textcolor{red}{25.61}&0.813 &\textcolor{blue}{30.49}&\textcolor{blue}{0.939}&\textcolor{blue}{27.82}&\textcolor{blue}{0.895}  \\
                % CSUD+Restormer (Ours)    &  \\
                % \rowcolor{blue!8}
                % CSUD+NeRD-Rain-S (Ours)   & \textcolor{red}{34.83} & \textcolor{red}{0.958} & \textcolor{blue}{24.02} & \textcolor{blue}{0.767} & \textcolor{blue}{34.06} & \textcolor{blue}{0.942} &22.75 &0.780 & \textcolor{blue}{31.99} & \textcolor{blue}{0.955} &  \textcolor{blue}{28.12}&\textcolor{blue}{0.901}\\
                % \rowcolor{blue!8}
            CSUD (Ours)    & \textcolor{red}{33.28} & \textcolor{red}{0.954} & \textcolor{red}{24.42} & \textcolor{red}{0.808} & \textcolor{red}{34.56} & \textcolor{red}{0.951} & 23.58& 0.795& \textcolor{red}{32.71} & \textcolor{red}{0.959} & \textcolor{red}{29.67} & \textcolor{red}{0.928} \\
        \bottomrule
      \end{tabular}
      }  
    \end{table*}

    \begin{figure*}[htb]
      \centering
      \setlength{\abovecaptionskip}{0.2cm}
      \includegraphics[width=1\textwidth]{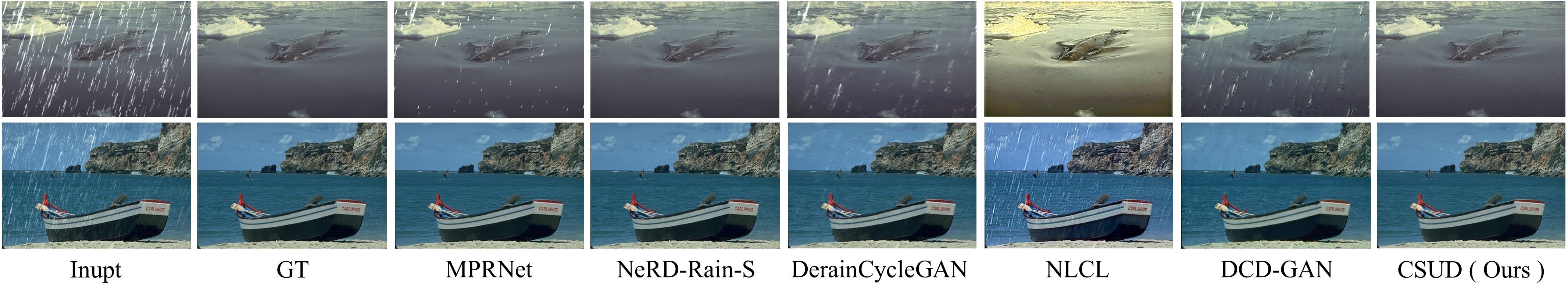}
      %\fbox{\rule[-.5cm]{0cm}{4cm} \rule[-.5cm]{4cm}{0cm}}
      \caption{Qualitative deraining performance comparisons on Rain100L \cite{Rain100L} and Rain12 \cite{Rain12} datasets. Our CSUD achieves competitive visual results comparable to the SOTA supervised method NeRD-Rain-S \cite{NeRD}.}
      % \vspace{-0.1cm}
      \label{fig.derain_results}
    \end{figure*}

    \section{Experiments}
    \label{Experiments}
    To demonstrate the effectiveness of our CSUD, following \cite{DerainCycleGAN, realrain1k}, we conduct experiments from the following two aspects: \textbf{(1) unsupervised deraining performance}, \textbf{(2) generalization performance}. In experiment (1), we train independent models for different datasets respectively, including both synthetic and real-world datasets. In experiment (2), we only train our model on the synthetic dataset Rain100L \cite{Rain100L}, and then test on various real-world and nighttime datasets.
    
    % \noindent
    \textbf{Datasets and Metrics.} We utilize 10 challenging benchmark datasets in our study, including synthetic datasets: Rain100L \cite{Rain100L}, Rain100H \cite{Rain100L}, Rain12 \cite{Rain12}, Rain800 \cite{Rain800}, real-world datasets: RealRain1K-L \cite{realrain1k}, RealRain1K-H \cite{realrain1k}, SPA-data \cite{SPANet}, RainDS \cite{RainDS}, Internet-Data \cite{SIRR}, and a nighttime dataset which is based on GTAV-NightRain dataset \cite{nightrain} and enlarged the rain density by \cite{nightrain2}, we name this nighttime dataset as Night-Rain in the following experiments. As for RainDS, we only use the real rain streaks subset for testing. Following existing methods \cite{restormer,DerainCycleGAN,DCDGAN}, we use PSNR and SSIM \cite{SSIM} to evaluate different methods. 
    
    % \textbf{Detailed descriptions of training and testing datasets employed and implementation details are provided in the ‘Appendix’.}
    
    % \noindent
    \textbf{Implementation Details.} As for derainer, we employ the CNN-based simple image restoration baseline NAFNet~\cite{NAFNet}. Our framework is implemented by PyTorch \cite{pytorch} with a GeForce RTX 3090 GPU. We adopt the Adam optimizer \cite{adam} to train our network for 200 epochs with the initial learning rate of $1e^{-4}$, followed by another 100 epochs with a learning rate of $1e^{-5}$. All training images are randomly cropped to 256 × 256 patches in an unpaired manner. The hyperparameters $\lambda_{1}$, $\lambda_{2}$, and $\lambda_{3}$ in Eq. \ref{eq12} are set to 1, 0.2 and 0.5, while $\alpha_{1}$ and $\alpha_{2}$ in Eq. \ref{eq13} are set to 10 and 5, respectively.
    % , and the batch size is set to 2
    % The hyperparameters of SSIM loss ($\lambda_{1}$), perceptual loss ($\lambda_{2}$), and SRLoss for derainer ($\lambda_{3}$) in Eq. \ref{eq12} are set to 1, 0.2 and 0.5 respectively, while CCLoss ($\alpha_{1}$) and SRLoss for generator ($\alpha_{2}$) in Eq. \ref{eq13} are set to 10 and 5 respectively.
    
    \begin{table*}[t]
    \scriptsize
      \caption{Quantitative generalization performance comparisons of different deraining methods. Red and blue colors indicate the 1st and 2nd ranks among the unsupervised methods. \textbf{S, U indicate supervised and unsupervised methods, respectively.} }
      % CSUD is only trained on Rain100L \cite{Rain100L}.
      % \resizebox{\textwidth}{!}{
      \label{table.generalization}
      \centering
      \scalebox{1.0}{
      \begin{tabular}{c|cccccccccc}
        \toprule
        \multicolumn{1}{c|}{Datasets} & \multicolumn{2}{c}{RealRain1K-L \cite{realrain1k}} & \multicolumn{2}{c}{RealRain1K-H \cite{realrain1k}} & \multicolumn{2}{c}{SPA-data \cite{SPANet}} & \multicolumn{2}{c}{RainDS \cite{RainDS}} & \multicolumn{2}{c}{Night-Rain \cite{nightrain2}}\\
        % \cmidrule(r){1-2}
        \midrule
        \multicolumn{1}{c|}{Metrics} & PSNR & SSIM & PSNR & SSIM & PSNR & SSIM & PSNR & SSIM & PSNR & SSIM  \\
        \midrule
        DerainNet \cite{DerainNet} (S)&25.40 &0.847&21.91& 0.742& 30.78 &0.912&21.35 &0.591&/&/\\  
         MPRNet \cite{MPRNet} (S)&26.75&0.899&23.58&0.840&33.00&0.942&23.24&0.656&26.13&0.817\\
                             NAFNet \cite{NAFNet} (S)  &28.11&0.915&24.36&0.848&33.46&0.941&22.48&0.625&26.15&0.816\\
                                          Restormer \cite{restormer} (S)     &26.29&0.883&23.13&0.851&32.62&0.936&22.96&0.648 &25.85 &0.808  \\
                                 PromptIR \cite{PromptIR} (S)       &28.14&0.915&24.28&0.851&33.50&0.940&22.48&0.625&26.37&0.815 \\
                                 NeRD-Rain-S \cite{NeRD} (S)  & 27.64 & 0.904 & 23.96& 0.831 & 33.03 & 0.933 & 22.37 & 0.622 & 26.38 & 0.810\\
        \midrule       
        DerainCycleGAN \cite{DerainCycleGAN} (U)    &28.25 & 0.899 & \textcolor{blue}{24.20}&0.821&\textcolor{blue}{33.53}&\textcolor{blue}{0.937}&18.75&0.460&\textcolor{blue}{25.85}&\textcolor{blue}{0.802}\\
        NLCL \cite{NLCL} (U)   &20.18&0.674&  19.81 &0.629&18.76&0.727&17.00&0.477 &22.01 &0.475\\
     DCD-GAN \cite{DCDGAN} (U) &  \textcolor{blue}{27.49}&\textcolor{blue}{0.912}&23.86& \textcolor{blue}{0.845}&31.55&0.930&\textcolor{blue}{22.13}&\textcolor{blue}{0.610}&25.54&0.798\\
     % CSUD+Restormer (Ours)    &  \\
     %  \rowcolor{blue!8}
     % CSUD+NeRD-Rain-S (Ours)   & \textcolor{blue}{29.02} & \textcolor{blue}{0.923} & \textcolor{blue}{25.23} & \textcolor{blue}{0.851} & 32.84 & 0.931  & \textcolor{blue}{22.26} & \textcolor{red}{0.618} & \textcolor{red}{26.19} & \textcolor{red}{0.807}\\
      % \rowcolor{blue!8}
     CSUD (Ours)    & \textcolor{red}{29.21} & \textcolor{red}{0.928} & \textcolor{red}{25.45} & \textcolor{red}{0.878} & \textcolor{red}{33.57} & \textcolor{red}{0.939} & \textcolor{red}{22.54} & \textcolor{red}{0.613} &\textcolor{red}{25.94}&\textcolor{red}{0.805}\\
        \bottomrule
      \end{tabular}
      }
      
    \end{table*}
    
    \begin{figure*}[t]
      \centering
      \setlength{\abovecaptionskip}{0.2cm}
      \includegraphics[width=1\textwidth]{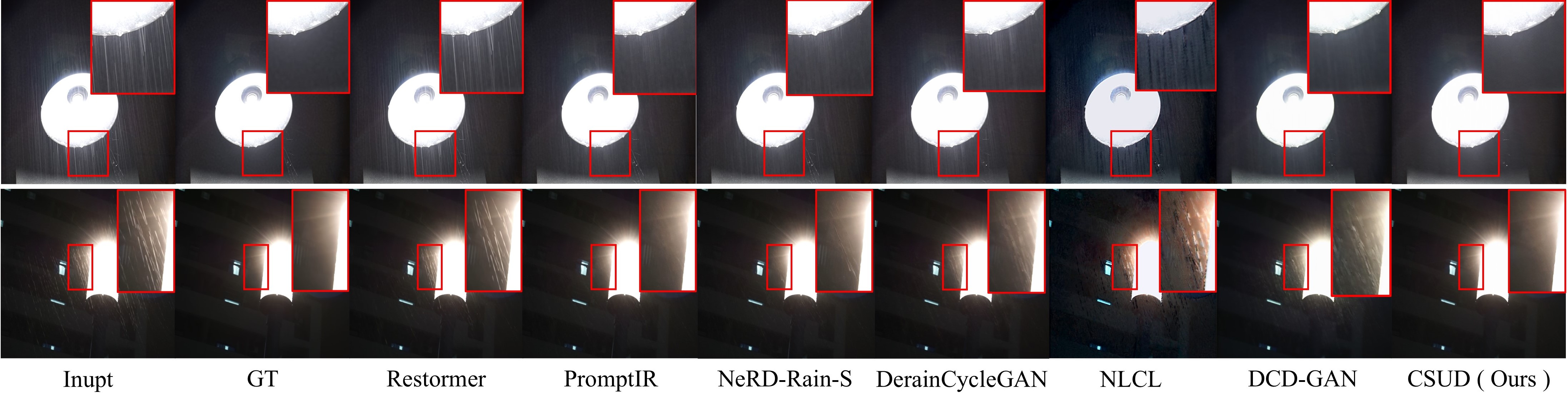}
      %\fbox{\rule[-.5cm]{0cm}{4cm} \rule[-.5cm]{4cm}{0cm}}
      \caption{Qualitative generalization results on RealRain1K-L \cite{realrain1k} and RealRain1K-H \cite{realrain1k} real-world datasets. Our CSUD achieves best visual results among both SOTA supervised and unsupervised methods in real-world scenes.}
        % \vspace{-0.1cm}
      \label{fig.Generalization_results}
    \end{figure*}

    \vspace{-0.1cm}
    \subsection{Unsupervised Deraining Results}
    \vspace{-0.1cm}
    We compare our method with unsupervised methods from the most recent years, including \cite{DerainCycleGAN, DCDGAN, NLCL, CycleGAN}. Noting that the related methods for comparison are limited, since only a few unsupervised deraining models have been proposed in the field. We also list some SOTA supervised methods, including \cite{DerainNet, DDN, SPANet, MPRNet, restormer, PromptIR, NAFNet, NeRD}. \cref{table.Unsupervisedderainingresults} presents the quantitative deraining results of different methods. It is evident that our CSUD significantly outperforms other unsupervised methods on most datasets. Compared to DCD-GAN\cite{DCDGAN}, our CSUD respectively achieves 1.46dB, 3.00dB, and 2.22dB improvement in PSNR on Rain100L, Rain12 and RealRain1K-L datasets. Additionally, our CSUD even achieves competitive results comparable to several supervised methods. As shown in \cref{fig.derain_results}, compared to other unsupervised methods, our CSUD achieves better deraining results while preserving more color and texture details of image background. 
            
            % The above results not only showcases the outstanding deraining performance but also the universality of CSUD. 

    % As shown in Figure \ref{fig.derain_results}, it can be seen that our CSUD achieves better results in removing rain streaks compared to other unsupervised methods. It is worth noting that there is a certain background color offset between the input and GT images of Rain800 dataset, though quantitative results are not the best, our CSUD preserves more color and texture details of image background while removing rain streaks. \textbf{Please refer to the ‘Appendix’ for more experimental results.}
    \vspace{-0.1cm}
    \subsection{Generalization Results}
    \vspace{-0.1cm}
    To validate the generalization capability of CSUD, we compare the results on 5 real-world datasets\cite{SPANet, SIRR, realrain1k, RainDS} and a nighttime dataset \cite{nightrain2} with SOTA supervised and unsupervised methods, including \cite{DerainNet, MPRNet, restormer, PromptIR, DerainCycleGAN, DCDGAN, NLCL, NAFNet, NeRD}. Noting that there are no ground truth images in the Internet-Data \cite{SIRR} dataset, thereby only visual results are presented. All methods are trained on synthetic datasets and tested on the unseen real-world and nighttime datasets. A prevailing perspective is that supervised methods usually outperform unsupervised methods. But as shown in \cref{table.generalization}, our CSUD almost comprehensively surpasses supervised models on the first 4 real-world datasets, while remaining competitive on the nighttime dataset. This strongly demonstrates the excellent generalization ability of our method.
    
    % The quantitative results intuitively demonstrate the outstanding generalization performance of our CSUD compared to other deraining models. 

\begin{figure*}[t]
      \centering
      \setlength{\abovecaptionskip}{0.2cm}
      \includegraphics[width=1\textwidth]{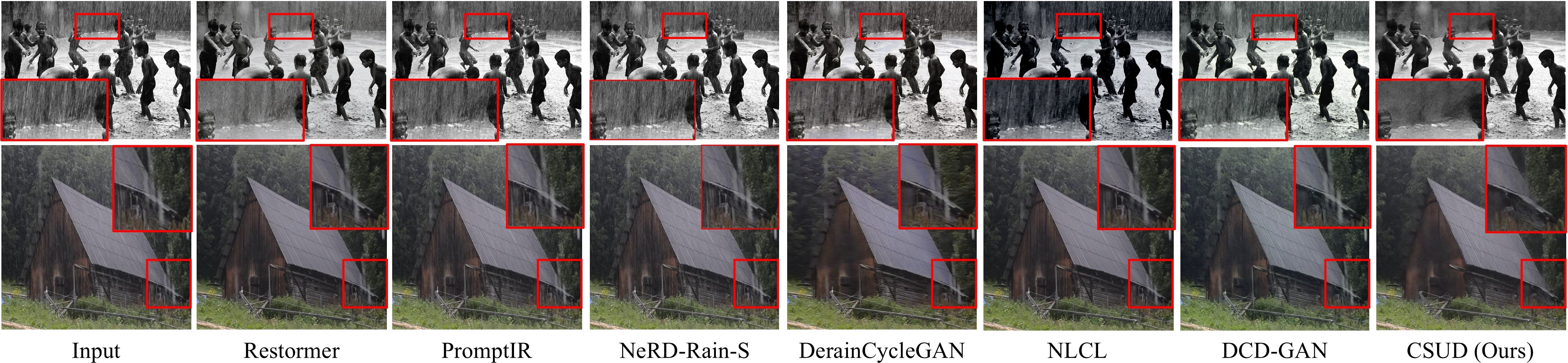}
      %\fbox{\rule[-.5cm]{0cm}{4cm} \rule[-.5cm]{4cm}{0cm}}
      \caption{Qualitative generalization results on Internet-Data \cite{SIRR} real-world dataset.}
        % \vspace{-0.1cm}
      \label{fig.Internet}
    \end{figure*}

    As shown in \cref{fig.Generalization_results} and \cref{fig.Internet}, the performance of some most advanced supervised methods like Restormer \cite{restormer} and NeRD-Rain-S \cite{NeRD} is extremely poor when directly applied to real-world scenarios. In contrast, our CSUD achieves better visual results in real world. It is worth mentioning again that, as shown in \cref{fig.realshot} in \cref{Introduction}, \textbf{our method also shows extremely superior deraining performance on the rainy images captured by ourselves in real scenarios,} which further demonstrates our excellent generalization capability. 
    % \textbf{Please refer to the Appendix for more experimental results.}
    
    % Furthermore, it can be clearly observed that our method not only removes rain streaks effectively but also better preserves the color and texture details of the background. 
    
    \subsection{Ablation Studies}
    \label{Ablation}
    
    \begin{table}[t]
    \scriptsize
    \caption{Effect of CCLoss and SR strategy on CSUD. }
      \label{table.AblationCCLossSRStrategy}
      \centering
    
        \begin{tabular}{c|c|cc|cc|cc}
        \toprule
         \multirow{2}{*}{CC}  & \multirow{2}{*}{SR} & \multicolumn{2}{c|}{Rain100L\cite{Rain100L}}  & \multicolumn{2}{c|}{Rain12\cite{Rain12}} & \multicolumn{2}{c}{RealRain1K-L\cite{realrain1k}}\\
       && PSNR & SSIM & PSNR & SSIM & PSNR & SSIM\\
        \midrule
        \XSolidBrush & \XSolidBrush &30.66 & 0.925  & 31.60 & 0.931 & 30.42 & 0.932 \\
        \Checkmark & \XSolidBrush &31.24 & 0.934 &32.62  & 0.948 & 31.14 & 0.943 \\
       \XSolidBrush & \Checkmark &32.69 & 0.949  & 33.43 & 0.939 & 31.86 & 0.950 \\
      %  \rowcolor{blue!8}
      \Checkmark & \Checkmark &\textbf{33.28} & \textbf{0.954 } & \textbf{34.56} & \textbf{0.951} & \textbf{32.71} &\textbf{0.959 } \\
        \bottomrule
        \end{tabular}    
    \end{table}
    
    % \noindent
    \textbf{Effect of CCLoss and SR Strategy on Deraining Performance.} To verify the effectiveness of CCLoss and SR strategy, we conduct ablation experiments on Rain100L \cite{Rain100L}, Rain12 \cite{Rain12} and RealRain1K-L \cite{realrain1k} datasets. As shown in Table \ref{table.AblationCCLossSRStrategy}, compared to the baseline, the effect on PSNR is improved by 1.38dB, 1.02dB, and 0.72dB among the above three datasets with CCLoss, respectively. And the effect is improved by 1.38dB, 1.83dB, and 1.44dB with SR strategy, respectively. Furthermore, when both CCLoss and SR strategy are introduced into the baseline, the PSNR on the three datasets is increased by 2.62dB, 2.96dB, and 2.29dB respectively. The above ablation results strongly verify the superiority and importance of CCLoss and SR strategy.
    
\begin{figure}[t]
      \centering
      \setlength{\abovecaptionskip}{0.2cm}
      \includegraphics[width=0.48\textwidth]{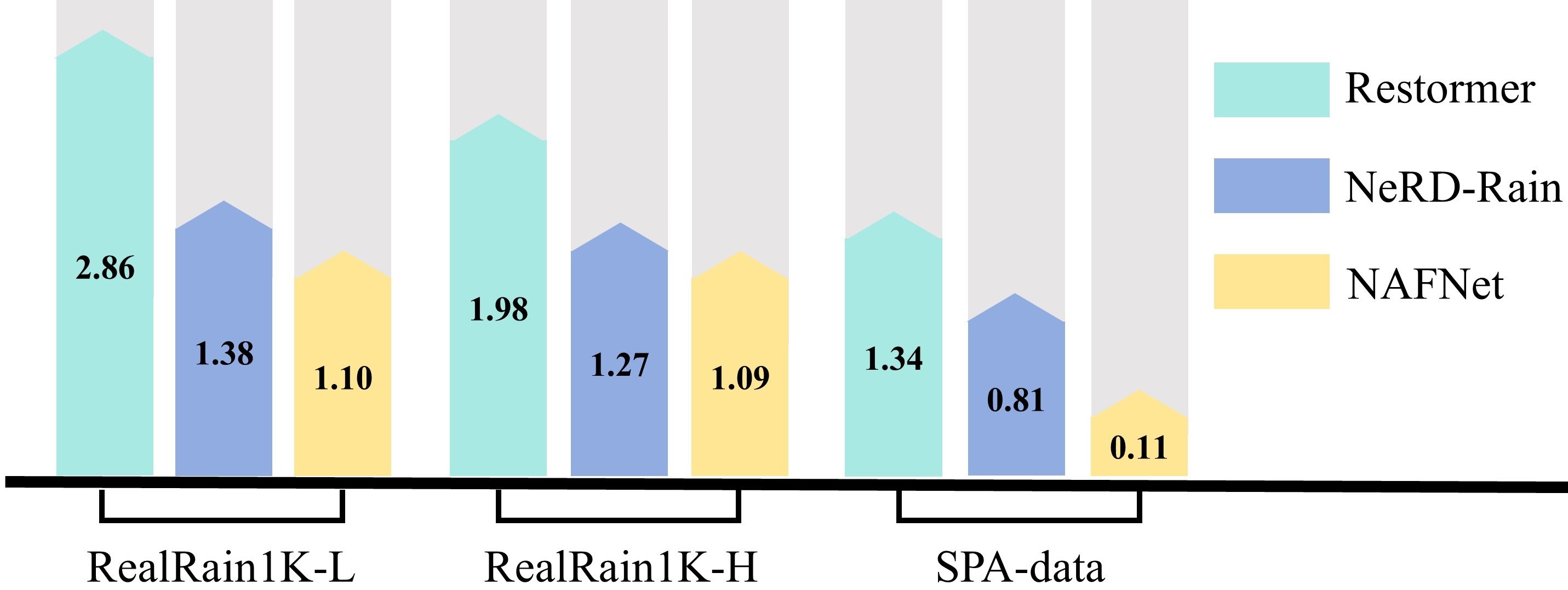}
      %\fbox{\rule[-.5cm]{0cm}{4cm} \rule[-.5cm]{4cm}{0cm}}
      \caption{PSNR improvement of different baselines with CSUD. We test the models' generalization performance on 3 real-world datasets out of domain compared with their supervised version.} 
      \label{fig.ablation2}
    \end{figure}

    % \noindent 
    \textbf{Effect of CSUD framework on Generalization Performance.} To validate the effectiveness of our CSUD in improving generalization performance and the universality and robustness of our method, we also conduct experiments incorporating CSUD framework on another 2 different baselines: NeRD-Rain-S~\cite{NeRD} and Restormer \cite{restormer}. \cref{fig.ablation2} shows the PSNR improvements of different CSUD based models compared to their supervised version. Notably, we train all the above models on Rain100L and test them on the 3 real-world datasets. This strongly demonstrates the effectiveness of CSUD in improving generalization performance. 
    
    % Figure \ref{fig.ablation2} (a) presents the improvement of PSNR on the three real-world datasets \cite{SPANet,realrain1k} with SR strategy, the effect on PSNR is increased by 1.30dB, 0.58dB, and 1.48dB respectively. 
    
    % \begin{figure}[htb]
    %   \centering
    %   \setlength{\abovecaptionskip}{0.cm}
    %   \includegraphics[width=0.5\textwidth]{fig/ablation sr.png}
    %   %\fbox{\rule[-.5cm]{0cm}{4cm} \rule[-.5cm]{4cm}{0cm}}
    %   \caption{Effect of SR strategy on generalization performance.}
    %   \label{fig.ablation}
    % \end{figure}
    
    % \begin{figure}[htb]
    %   \centering
    %   \setlength{\abovecaptionskip}{0.cm}
    %   \includegraphics[width=0.45\textwidth]{author-kit-CVPR2025-v3.1-latex-/fig/ablation.jpg}
    %   %\fbox{\rule[-.5cm]{0cm}{4cm} \rule[-.5cm]{4cm}{0cm}}
    %   \caption{Effect of SR strategy on generalization performance.}
    %   \label{fig.ablation2}
    % \end{figure}
    
    \textbf{Effect of Different Training Datasets on Generalization Performance.} To further validate the generalization performance of CSUD, we employ 4 different CSUD models which are respectively trained on 4 synthetic and real-world datasets with different rain distributions, and test them on 3 real-world datasets. As shown in \cref{table.generalization22}, in most cases, CSUD can achieve strong generalization performance, demonstrating the robustness of our approach. Due to the significant differences between rain streaks in Rain100H and real world, its generalization capability is the worst. Additionally, the generalization performance when training on Rain100L is comparable to that when training on real-world datasets, demonstrating our strong generalization ability.
    % However, thanks to the superiority of our framework, it still maintains relatively high generalization performance.
    
    \begin{table}[t]
    % \scriptsize
      \caption{Effect of different training datasets on generalization performance. We test 4 different CSUD models which are trained on 4 different datasets on 3 real-world datasets out of domain.}
      \centering
      % CSUD is only trained on Rain100L \cite{Rain100L}.
      % \resizebox{\textwidth}{!}{
      \scalebox{0.7}{
      \label{table.generalization22}
      \centering
      \begin{tabular}{c|cccccc}
        \toprule 
        \multirow{2}{*}{Train datasets} & \multicolumn{2}{c}{SPA-data \cite{SPANet}} & \multicolumn{2}{c}{Night-Rain \cite{nightrain2}} & \multicolumn{2}{c}{RainDS \cite{RainDS}} \\
        & PSNR & SSIM & PSNR & SSIM & PSNR & SSIM\\
        % \cmidrule(r){1-2}
        \midrule
        Rain100L \cite{Rain100L} &33.57& 0.939 &25.94 &0.805 & 22.54 &0.613  \\  
        % Rain100H \cite{Rain100L}  &30.05 & 0.898 & 24.71& 0.714&22.10 & 0.596\\
         Rain100H \cite{Rain100L}  &29.61 & 0.889 & 24.53& 0.702&21.85 & 0.600\\
        RealRain1K-L \cite{realrain1k} &32.95&0.937 & 28.14 & 0.835 &22.90 & 0.630\\
        RealRain1K-H \cite{realrain1k}  & 29.25 & 0.897 & 26.31 & 0.787 & 22.30 &0.564\\
        \bottomrule
      \end{tabular}
      }
    \end{table}

    % \noindent
    
    \textbf{Effect of the Weight Coefficients of Each Loss Function.} We conduct extensive ablation experiments on the weight coefficients of perceptual loss in~\cref{eq12}, $L_{CC}$ and $L_{SR-G}$ in~\cref{eq13}. The results are shown in~\cref{table.coefficients}. Since $L_{SR-Der}$ and $L_{SR-G}$ are highly correlated, to ensure that the losses are on the same scale, the coefficient for $L_{SR-Der}$ is set to 0.1 times that of $L_{SR-G}$, and $L_{SR}$ in~\cref{table.coefficients} indicates $L_{SR-G}$. Based on the experimental results, aside from our selected optimal combination, the deraining and generalization performance of our method remain at a high level under most other weight combinations, demonstrating the robustness of our CSUD framework.
    
    \textbf{More detailed explanations of our method, experimental details and analysis can be found in the Appendix.}
    
    \begin{table}[htb]
    \scriptsize
      \caption{Ablation experiments on the weight coefficients of each loss function. All models in the table are trained on Rain100L.}
      % CSUD is only trained on Rain100L \cite{Rain100L}.
      % \resizebox{\textwidth}{!}{
      \label{table.coefficients}
      \scalebox{0.87}{
      \centering
      \begin{tabular}{c|c|c|cc|cc|cc}
        \toprule
        $L_{per}$& $L_{SR}$& $L_{CC}$& \multicolumn{2}{c|}{Rain100L \cite{Rain100L}} & \multicolumn{2}{c|}{SPA-data \cite{SPANet}} & \multicolumn{2}{c}{RealRain1K-L \cite{realrain1k}}    \\
        % \cmidrule(r){1-2}
        % \midrule
        ($\lambda_{2})$&$(\alpha_{2})$&($\alpha_{1}$)& PSNR & SSIM & PSNR & SSIM & PSNR & SSIM   \\
        \midrule
        % NLCL_{ResBlock}&20.50&0.719& 18.76&0.727 & 20.18&0.674 \\  
        0.2 & 5.0& 0.1&30.69&0.932&32.13&0.929&28.21&0.915 \\
        0.2 & 5.0& 1.0&33.15&0.951 &33.50&0.938 &\textbf{29.75}&\textbf{0.928}\\
        % 0.2 & 5.0& 50 &32.23&0.946 &32.31&0.926 & 27.73&0.912 \\
        0.2 &1.0 &10 &32.84&0.946&33.49&0.935 &29.10&0.918 \\
        0.2 &10 &10 &32.90&0.951&33.40&0.933 &29.06&0.920 \\
        0 &5.0 &10 &32.18&0.939&32.69&0.936 & 28.24&0.916 \\
        1.0 &5.0 &10 &32.72&0.947 &33.05&0.930 & 28.70&0.918\\
        % \rowcolor{blue!8}
        0.2&5.0 &10&\textbf{33.28}&\textbf{0.954}&\textbf{33.57}&\textbf{0.939}&29.21&\textbf{0.928}\\
        \bottomrule
      \end{tabular}
      }
    \end{table}

    \section{Conclusion}
    In this paper, we propose a novel unsupervised framework for image deraining to address the lack of paired data and the poor generalization in real-world scenarios. We first propose a channel consistency loss by introducing the channel consistency prior into the optimization process of the generator, ensuring that more background details are preserved while generating pseudo rainy images. In addition, a self-reconstruction strategy is proposed to alleviate the redundant information transfer problem and further improve the deraining performance and the generalization capability of our method. Extensive experiments on multiple synthetic and real-world datasets verify that our method achieves excellent deraining performance with strong generalization capability across various real-world rainy scenarios.
    \\
    
    % 
    % \clearpage
    \vspace{-0.08cm}
    \noindent
    \textbf{Acknowledgement.} This work was supported by the National Natural Science Foundation of China under Grant 62171304 and partly by the Natural Science Foundation of Sichuan Province under Grant 2024NSFSC1423, the TCL Science and Technology Innovation Fund under grant 25JZH008, the Young Faculty Technology Innovation Capacity Enhancement Program of Sichuan University under Grant 2024SCUQJTX025, and the Beijing Natural Science Foundation L221011.
    \newpage

{
    \small
    \bibliographystyle{ieeenat_fullname}
    \bibliography{main}
}

\clearpage
\setcounter{page}{1}
\setcounter{table}{0}
\setcounter{figure}{0}
\setcounter{section}{0}
\maketitlesupplementary
% \section*{Appendix}

\begin{figure}[t]
  \centering
  \setlength{\abovecaptionskip}{0.1cm}
  \includegraphics[width=0.47\textwidth]{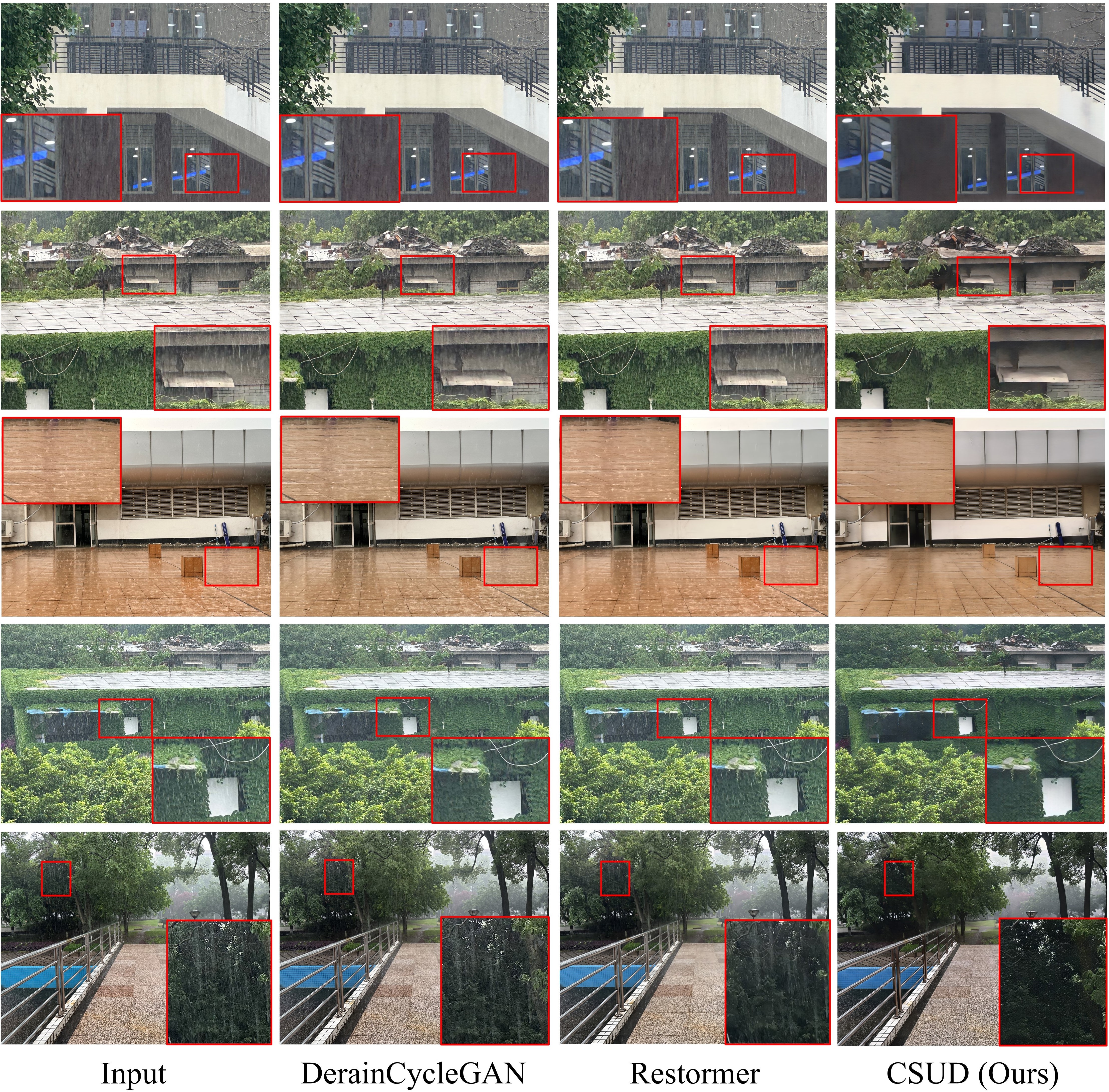}
  %\fbox{\rule[-.5cm]{0cm}{4cm} \rule[-.5cm]{4cm}{0cm}}
  \caption{Deraining results on the real rainy images captured by ourselves in real-world scenarios. Compared with the supervised method Restormer \cite{restormer} and the unsupervised method DerainCycleGAN \cite{DerainCycleGAN}, our CSUD exhibits extremely strong generalization capability and achieves the best visual results.}
  \label{fig.SM_realshot}
\end{figure}

In this paper, we propose a novel \textbf{C}hannel Consistency Prior and \textbf{S}elf-Reconstruction Strategy based \textbf{U}nsupervised Image \textbf{D}eraining framework, \textbf{CSUD}, to address the lack of paired data and the poor generalization in real-world scenarios. Extensive experiments on multiple synthetic and real-world datasets verify that our method achieves excellent deraining performance. What's more, as shown in \cref{fig.SM_realshot}, our CSUD achieves the most promising performance on the real-world captured images compared to both supervised and unsupervised deraining methods, which demonstrates strong generalization capability of our CSUD. This supplementary material mainly includes the following contents:
\begin{itemize}
\item More detailed explanations of the proposed channel consistency prior and self-reconstruction strategy;
\item The specific structure of certain used networks;
\item More implementation details of the experiments mentioned in the main document; 
\item Additional experiment results and analysis;
\item Discussion and limitations of our method.
\end{itemize}

%%%%%%%%% BODY TEXT
\section{More Explanations of Channel Consistency Prior}
\begin{figure*}[p]
  \centering
  \setlength{\abovecaptionskip}{0.cm}
  \includegraphics[width=0.9\textwidth]{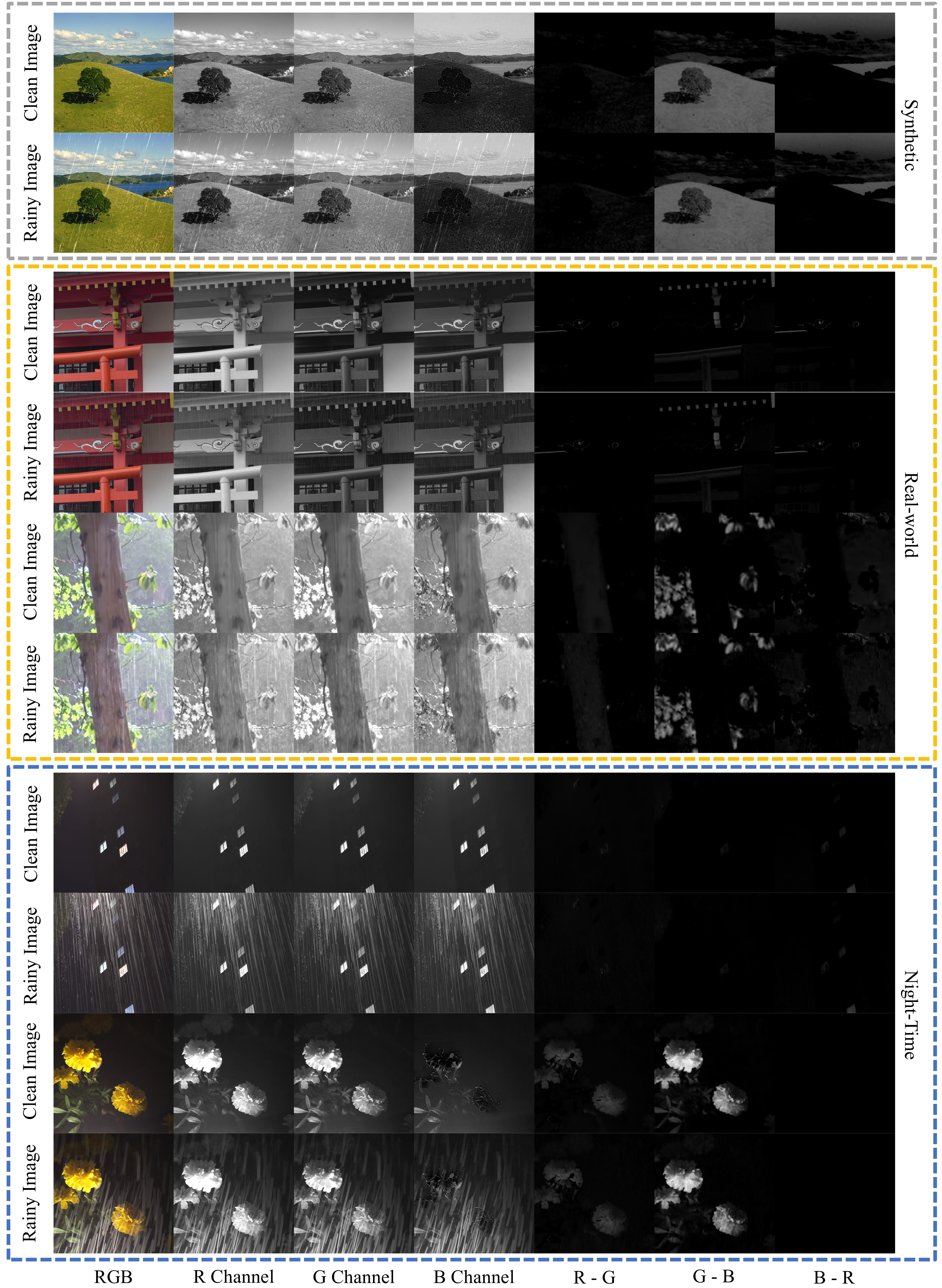}
  %\fbox{\rule[-.5cm]{0cm}{4cm} \rule[-.5cm]{4cm}{0cm}}
  \caption{Visualization of channel consistency prior in rainy images. From top to bottom, the 3 sets of images are synthetic, real-world, and real-world nighttime images, respectively. The first column presents the clean and rainy RGB images; the second, third and fourth columns present their R, G, B channels, respectively; the fifth, sixth, and seventh columns present the cycle subtractions of R, G, and B channels of clean and rainy images, respectively.}
   \label{fig.CPP_Details}
\end{figure*}

% \section{CCP in real-world rainy images and real-world nighttime rainy images}
In the main document, we have provided a detailed introduction to the channel consistency prior (CCP) of rain streaks: in RGB rainy images, most rain streaks tend to be consistent across the R, G, and B channels, and the cycle subtractions of R, G, and B channels in rainy images will almost contain no rain streaks and should be close to the cycle subtractions in clean images. To present the conclusions derived from CCP more clearly and intuitively, we provide a more detailed visualization of the CCP in rainy images along with the corresponding clean images in \cref{fig.CPP_Details}, including synthetic rain, real-world rain, and real-world nighttime rain. We can clearly observe that while there are significant differences in the background parts among the three channels of rainy images, the rain streaks tend to be consistent, and the cycle subtractions of R, G, and B channels in rainy images tend to be consistent with that of their corresponding clean images.  

\begin{figure*}[t]
  \centering
  \includegraphics[width=0.9\textwidth]{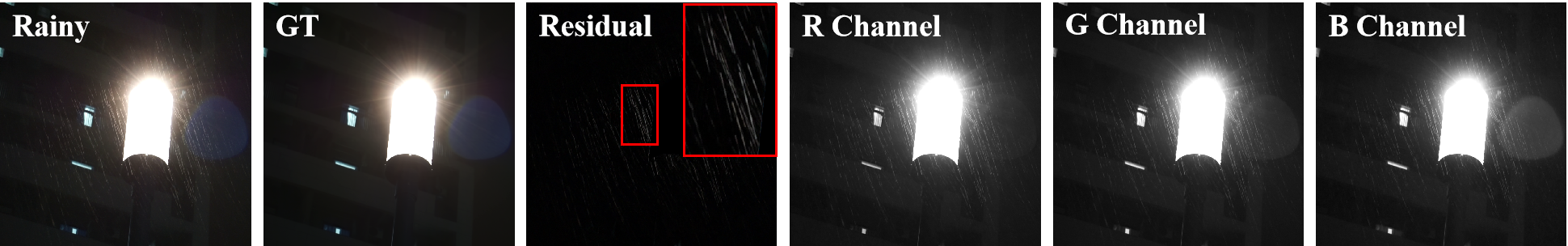}
  %\fbox{\rule[-.5cm]{0cm}{4cm} \rule[-.5cm]{4cm}{0cm}}
  % \caption{\normalsize CCP in night time.} 
  % \captionsetup{font={footnotesize}}
    \caption{Visualization of channel consistency prior under the nighttime artificial light scenario. }
  \label{fig.rebuttal_ccp_nighttime}
\end{figure*}

Currently, the rain in an image is typically modeled using the linear superposition model:  $Y = X + R $, where $ Y $ is the rainy image, $ R $ is the rain layer, and $ X $ is the image background. Most image deraining researches \cite{DerainNet, DerainCycleGAN, DCDGAN, DIDMDN,NeRD,DRSformer,SEIDNet} focuses on the removal of the rain layer $ R $ in images. One of the reasons why many unsupervised image deraining models perform poorly is that, while they remove rain layer, they struggle to maintain the integrity of other background information in the image. In this work, we also mainly focus on removing rain layer $ R $ from rainy images. During training, the derainer learns the mapping between pseudo-rainy images and real clean images $ X $. If the pseudo-rainy images maintain the main background information of $ X $, differing from clean images only by the presence of the rain streaks, the derainer will learn the optimal mapping relationship. Thus, we introduce the Channel Consistency Loss (CCLoss) to constrain the generator $G$ to produce pseudo-rainy images that retain most of the image content consistent with the original clean images $ X $ aside from the rain streaks. Through the constrain of CCLoss, the derainer can better learn the desired mapping.

In addition, we separately analyze the special cases which include nighttime artificial light sources. As shown in \cref{fig.rebuttal_ccp_nighttime}, we provide a visualization of CCP \textbf{under the nighttime artificial light scenario}. While the rain streaks in rainy image appear yellowish due to interference from artificial light, the rain streaks in the residual image do not exhibit such yellowish tones, and most rain streaks still conform to CCP among R,G,B channels. According to the linear superposition model, rainy images can be understood as the superposition of the rain layer and the clean background. Therefore, the yellowish hue in rain streaks is the color of the background introduced by artificial light. Futhermore, in Fig.7, Tab.1, and Tab.2 of the main text, we test our model on real-world datasets RealRain1K-L and RealRain1K-H which \textbf{include numerous nighttime artificial light sources}, and our CSUD achieves remarkable results. 

Certainly, there may be some rain streaks that do not adhere to the CCP, but this does not significantly affect the network's performance. This is because, besides CCP, the SR strategy and overall unsupervised framework with the three additional constraints based on CycleGAN \cite{CycleGAN} can also ensure the accurate and effective generation of pseudo-paired rainy images, as well as the deraining and generalization performance of CSUD. The CCLoss serves as a further auxiliary enhancement to the overall framework, aiming to transfer rain streaks while preserving more background details as much as possible, which makes our framework more robust. It is the combined effort of our unsupervised framework, CCLoss, and the SR strategy that achieves the outstanding unsupervised deraining performance of CSUD.

% As long as the majority of rain streaks in rainy images conform to CCP, it can serve as a constraint during training and enhance the performance. 

\section{More Explanation of Self-Reconstruction Strategy}
\begin{figure}[t]
  \centering
  \setlength{\abovecaptionskip}{0.2cm}
  \includegraphics[width=0.45\textwidth]{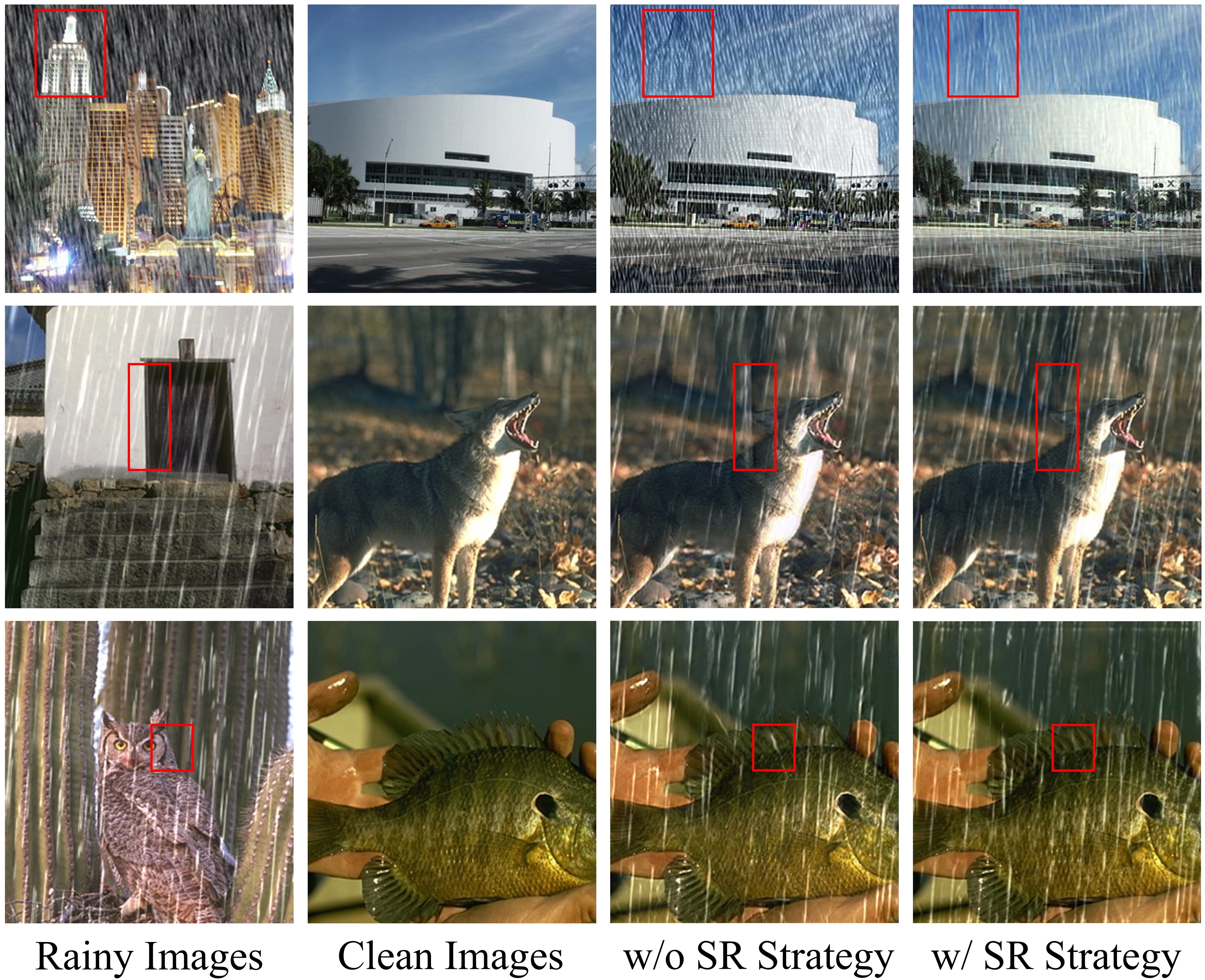}
  %\fbox{\rule[-.5cm]{0cm}{4cm} \rule[-.5cm]{4cm}{0cm}}
  \caption{The effects of SR strategy on the performance of generator. The first and second columns present the input rainy and clean images of the generator, respectively; the third column presents pseudo rainy images generated by the generator without SR strategy; the fourth column presents pseudo rainy images generated by the generator with SR strategy.}
  \label{fig.RITP_MORE}
\end{figure}

As described in main document, $x_{der}$ restored by derainer $Der$ is not adopted in SRLoss for generator $L_{SR-G}$. However, $x_{der}$ can also be used in $L_{SR-G}$ to constrain the training process of generator $G$, but we do not add it in the final implementation in order to make the SR loss for generator $L_{SR-G}$ and SR loss for derainer $L_{SR-Der}$ on the same scale and reduce the calculation of the training process, which has little effect on the performance of the network. Additionally, as shown in \cref{fig.RITP_MORE}, we present more visualizations of the effects of SR strategy on the performance of generator $G$. With SR strategy, the generator effectively alleviates the redundant information transfer problem, ensuring that higher-quality pseudo rainy images are generated.

\section{Detailed Network Structures.}

As described in the main document, the proposed CSUD mainly consists of a derainer, a generator, and a discriminator. To balance the performance and computational complexity, we adopt the simple CNN-based image restoration baseline NAFNet \cite{NAFNet} (the version of width32) as the default derainer. 
% The structures of the generator and the discriminator will be detailed below.

\begin{figure}[t]
  \centering
  \setlength{\abovecaptionskip}{0.2cm}
  \includegraphics[width=0.45\textwidth]{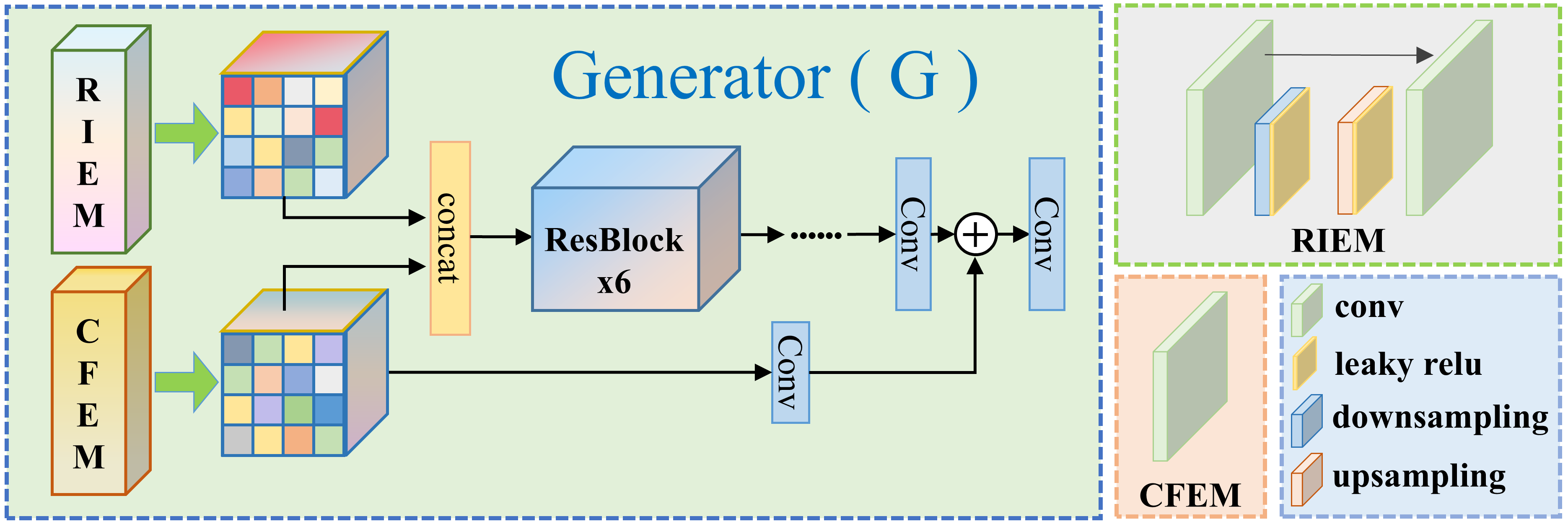}
  %\fbox{\rule[-.5cm]{0cm}{4cm} \rule[-.5cm]{4cm}{0cm}}
  \caption{Detailed network structures of the generator.}
  \label{fig.generator}
\end{figure}

\textbf{Architecture of the Generator.} The architecture of the generator used in our network is shown in \cref{fig.generator}, it consists of a clean feature extraction module (CFEM), a rain information extraction module (RIEM) and 6 residual blocks. The RIEM is based on a U-Net architecture, comprising a downsampling layer followed by an upsampling layer, while the CFEM simply utilize a convolutional layer. Specifically, the two convolutional layers in RIEM (denoted as "conv") have input channels, output channels, kernel size, stride, and padding settings of [3, 64, 7, 1, 3] and [64, 64, 7, 1, 3], respectively. The downsampling layer in RIEM is a convolutional layer with input channels, output channels, kernel size, stride, and padding set to [64, 128, 4, 2, 1]. The upsampling layer in RIEM uses a 'bilinear' interpolation, followed by a convolutional layer with input channels, output channels, kernel size, stride, and padding set to [128, 64, 3, 1, 1]. The convolutional layer in CFEM has input channels, output channels, kernel size, stride, and padding set to [3, 64, 7, 1, 3]. Each residual block consists of two 3 \( \times  \) 3 convolution layers with ReLU activation function. The generator learns the rain characteristics of rainy images to guide the synthesis of clean images towards rainy ones, providing ample pseudo rainy images paired with the clean ones for the derainer.

\begin{figure}[t]
  \centering
  \setlength{\abovecaptionskip}{0.2cm}
  \includegraphics[width=0.42\textwidth]{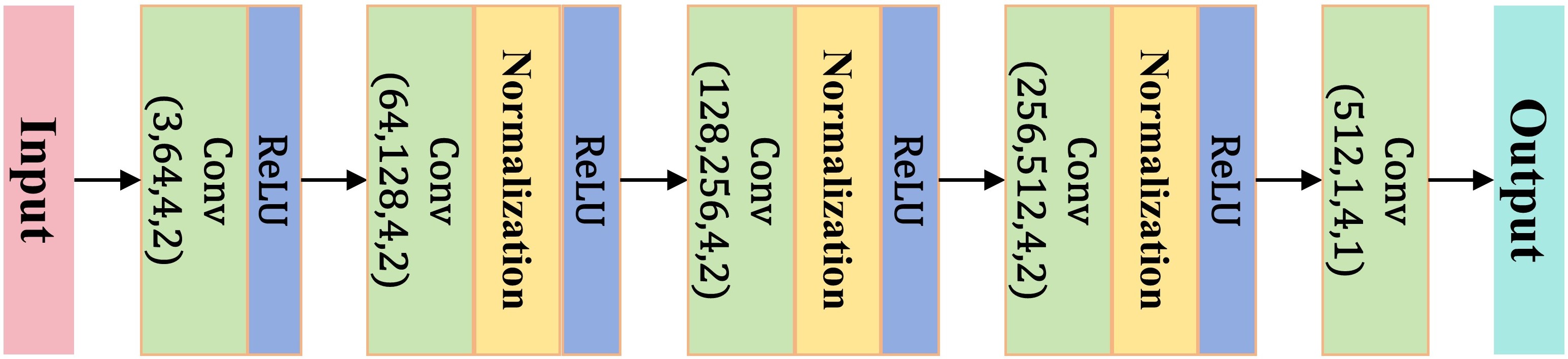}
  %\fbox{\rule[-.5cm]{0cm}{4cm} \rule[-.5cm]{4cm}{0cm}}
  \caption{Detailed network structures of the discriminator.}
  \label{fig.discriminator}
\end{figure}

\textbf{Architecture of the Discriminator.}
In our network, we use a Patch-GAN \cite{PatchGAN} discriminator, as shown in Fig. \ref{fig.discriminator}. The discriminator is starting with a 4 \( \times  \) 4 convolution layer with ReLU activation function, followed by three intermediate layers, each of which consists of instance normalization between the convolution layer and the activation function, and ending with a 4 \( \times  \) 4 convolution layer with a stride of 1. 
% The adversarial training between the discriminator and the generator ensures that the synthesized rainy images are close to real ones.

\section{More Explanation of Experiment Setting}
As described in the main document, following \cite{DerainCycleGAN, realrain1k}, we conduct experiments from two aspects: (1) unsupervised deraining performance and (2) generalization performance. In experiment (2), we only train our model on the synthetic dataset Rain100L \cite{Rain100L}, and then test on various real-world and nighttime datasets. This experimental setup is designed to better demonstrate the strong cross-domain generalization ability of our CSUD framework, specifically its deraining performance when faced with various rain streak distributions, rather than suggesting that our CSUD should be trained exclusively on synthetic datasets for optimal performance. Our CSUD can also be trained on unpaired real-world datasets and achieve better performance. In the main document, we have conducted experiments on the real-world RealRain-1k-L and RealRain-1k-H datasets according to the experiment (1) settings. Trained on unpaired real-world datasets, CSUD achieves better results on the two real-world datasets, even surpassing some classic supervised methods.
% Additionally, in ablation studies of main document, we further demonstrate that our CSUD also exhibits excellent generalization ability when trained on different real-world and synthetic datasets.

% This setup follows the practices of many excellent deraining works \cite{DCDGAN,DerainCycleGAN,realrain1k}, to ensure fair comparison with these methods.

% \renewcommand{\thesection}{A\arabic{section}}
\section{Experiment Details}

\textbf{Datasets.} Detailed descriptions of the datasets employed are provided in \cref{table.datasets}. In experiment (1), we use the corresponding different training sets to train independent models for Rain100L \cite{Rain100L}, Rain100H \cite{Rain100L}, Rain800 \cite{Rain800}, RealRain1K-L\cite{realrain1k}, and RealRain1K-H \cite{realrain1k} test sets respectively. Notably, we utilize the model trained on Rain100L to test on Rain12 \cite{Rain12} dataset. As for experiment (2), we only train our model on Rain100L \cite{Rain100L}, and then test on the 6 real-world and night-time test sets, including RealRain1K-L\cite{realrain1k}, RealRain1K-H \cite{realrain1k}, SPA-data \cite{SPANet}, RainDS \cite{RainDS}, Internet-Data \cite{SIRR}, and Night-Rain \cite{nightrain2}. It is worth noting that RainDS includes multiple subsets, including synthetic and real subsets, with the two subsets further divided into rain streaks, rain drops, and a mixture of rain streaks and drops. Since our method focuses on removing rain streaks and experiment (2) is to evaluate generalization performance on real-world and night-time test sets, so we only select the rain streaks subset from the real RainDS subset for testing. All other comparison models are also tested on this subset.

\begin{table}[t]
  \scriptsize
  \caption{Detailed description of the datasets utilized.}
  \resizebox{0.48\textwidth}{!}{
  \label{table.datasets}
  \centering
  \begin{tabular}{c|ccccc}
    \toprule
    Datasets & Rain100L \cite{Rain100L}& Rain100H \cite{Rain100L}& Rain12 \cite{Rain12}& Rain800 \cite{Rain800} &RealRain1K-L \cite{realrain1k}   \\
    % \cmidrule(r){1-2}
    \midrule
    Train & 200 & 200 & 0 & 700 & 784\\
    Test & 100 & 100 & 12 & 100 & 224\\
    Rain Type & Synthetic & Synthetic & Synthetic & Synthetic & Real-world\\
    \bottomrule
    \toprule
    Datasets & RealRain1K-H \cite{realrain1k} & SPA-data \cite{SPANet} & RainDS \cite{RainDS} &Internet-Data \cite{SIRR}  & Night-Rain \cite{nightrain2} \\
    % \cmidrule(r){1-2}
    \midrule
    Train   & 784 &638,492&150&0& 5000\\
    Test   & 224 &1000&98&147& 500\\
    Rain Type   & Real-world&Real-world&Real-world&Real-world& Night-Time \\
    \bottomrule
  \end{tabular}
  }
\end{table}

\begin{figure*}[t]
  \centering
  \setlength{\abovecaptionskip}{0.2cm}
  \includegraphics[width=0.99\textwidth]{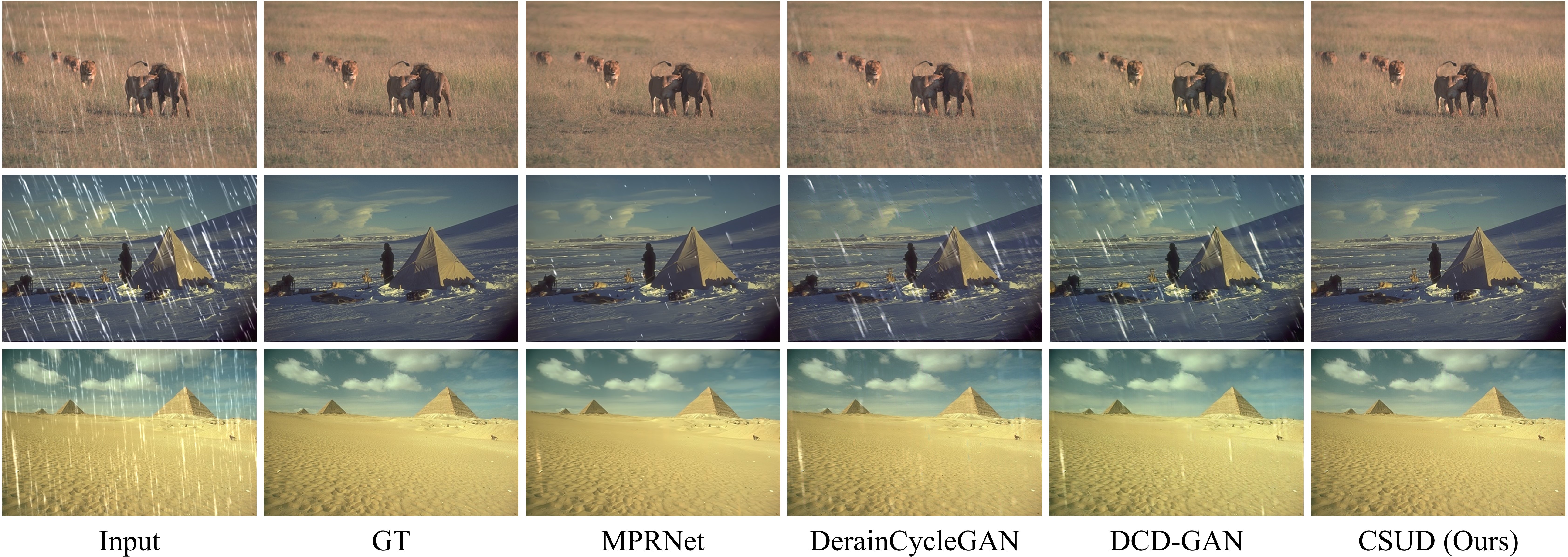}
  %\fbox{\rule[-.5cm]{0cm}{4cm} \rule[-.5cm]{4cm}{0cm}}
  \caption{Qualitative deraining results on Rain100L \cite{Rain100L} dataset.}
  \label{fig.SM_rain100l_results}
\end{figure*}

\begin{figure*}[t]
  \centering
  \setlength{\abovecaptionskip}{0.2cm}
  \includegraphics[width=0.99\textwidth]{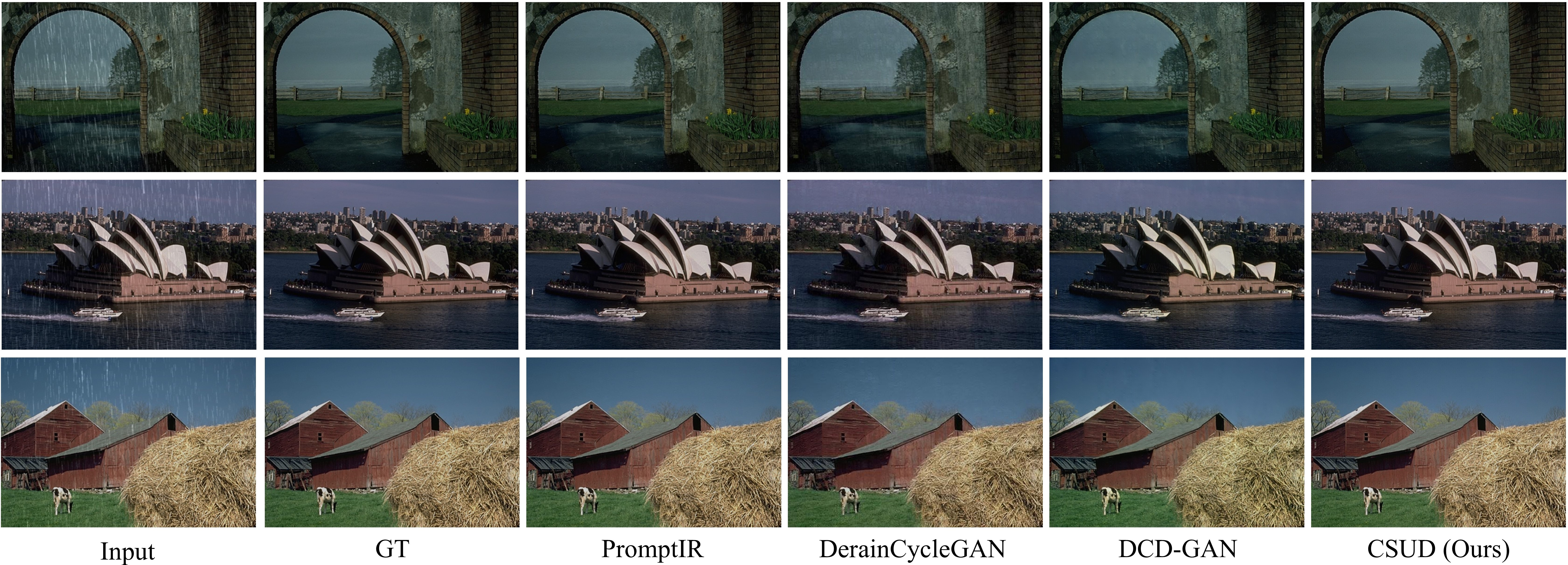}
  %\fbox{\rule[-.5cm]{0cm}{4cm} \rule[-.5cm]{4cm}{0cm}}
  \caption{Qualitative deraining results on Rain12 \cite{Rain12} dataset.}
  \label{fig.SM_Rain12}
\end{figure*}

\begin{figure*}[t]
  \centering
  \setlength{\abovecaptionskip}{0.2cm}
  \includegraphics[width=0.99\textwidth]{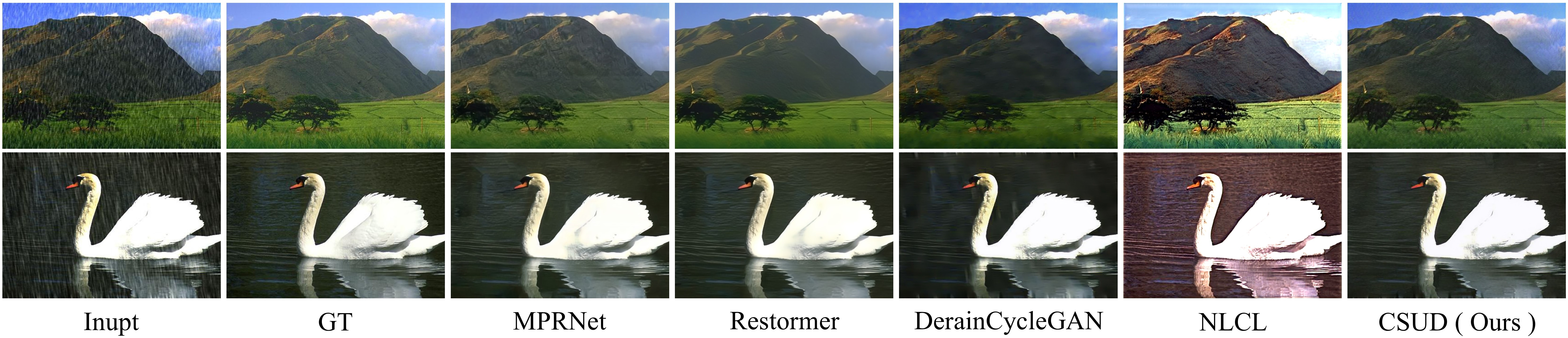}
  %\fbox{\rule[-.5cm]{0cm}{4cm} \rule[-.5cm]{4cm}{0cm}}
  \caption{Qualitative deraining results on Rain800 \cite{Rain800} dataset.}
  \label{fig.SM_Rain800}
\end{figure*}

\begin{table*}[t]
\scriptsize
  \caption{Quantitative perceptual quality comparisons of different deraining baselines with or without our methods. }
  % CSUD is only trained on Rain100L \cite{Rain100L}.
  % \resizebox{\textwidth}{!}{
  \label{table.generalization_perceptual}
  \centering
  \scalebox{0.9}{
  \begin{tabular}{c|ccc}
    \toprule
    \multicolumn{1}{c|}{Datasets} & \multicolumn{1}{c}{RealRain1K-L \cite{realrain1k}} & \multicolumn{1}{c}{RealRain1K-H \cite{realrain1k}} & \multicolumn{1}{c}{SPA-data \cite{SPANet}} \\
    % \cmidrule(r){1-2}
    \midrule
    \multicolumn{1}{c|}{Metrics} & LPIPS $\downarrow$ / DISTS $\downarrow$/ NIQE $\downarrow$& LPIPS $\downarrow$ / DISTS $\downarrow$/ NIQE $\downarrow$& LPIPS $\downarrow$ / DISTS $\downarrow$/ NIQE $\downarrow$\\
    \midrule 
     MPRNet (Supervised) &0.355 / 0.279 / \textbf{8.872}& 0.424 / 0.314 / \textbf{8.231}&0.159 / 0.125 / 7.946\\ 
    %  \rowcolor{blue!8}
     MPRNet + CSUD (Unsupervised) & \textbf{0.228} / \textbf{0.213} / 9.874 & \textbf{0.271} / \textbf{0.241} / 9.452 & \textbf{0.151} / \textbf{0.124} / \textbf{7.532}\\
     \midrule 
    % Restormer \cite{restormer} (S)  &  \\
    NeRD-Rain-S (Supervised)  &0.341 / \textbf{0.298} / 7.132& 0.445 / 0.339 / 6.679 &0.167 / 0.131 / 7.151\\
    %  \rowcolor{blue!8}
    NeRD-Rain-S + CSUD (Unsupervised)  & \textbf{0.336} / \textbf{0.298} / \textbf{7.104}& \textbf{0.441} / \textbf{0.338} / \textbf{6.489} & \textbf{0.160 }/ \textbf{0.130} / \textbf{7.079}  \\
    \midrule       
  NAFNet (Supervised)  &0.308 / 0.285 / \textbf{7.150} & 0.416 / 0.328 / \textbf{6.722}&0.152 / 0.124 / \textbf{7.108}\\
  % \rowcolor{blue!8}
NAFNet + CSUD (Unsupervised)    & \textbf{0.258} / \textbf{0.257} / 8.103&\textbf{0.345} / \textbf{0.295} / 7.622&\textbf{0.141} /\textbf{ 0.120 }/ 7.369\\
    \bottomrule
  \end{tabular}
  }
  
\end{table*}

\textbf{Implementation Details}. Our framework is implemented by PyTorch \cite{pytorch} with a GeForce RTX 3090 GPU. For training, we adopt the Adam optimizer \cite{adam} ($\beta_{1}$ = 0.9, $\beta_{2}$ = 0.999) to train our network. We train the framework for 200 epochs with the initial learning rate of $1e^{-4}$, followed by another 100 epochs with a learning rate of $1e^{-5}$. All training images are randomly cropped to 256 × 256 patches in an unpaired learning manner, and the batch size is set to 2. The hyperparameters of SSIM loss ($\lambda_{1}$), perceptual loss ($\lambda_{2}$), and SRLoss for derainer ($\lambda_{3}$) are set to 1, 0.2 and 0.5 respectively, while CCLoss ($\alpha_{1}$) and SRLoss for generator ($\alpha_{2}$) are set to 10 and 5 respectively. Notably, we add perceptual loss to our framework is not to improve perceptual quality of our results, but to enhance the stability of the unsupervised training process. Because we find that only using L1 loss as the constraint of the derainer will collapse in the middle of the training process, which is caused by the difficulty and instability of the GAN manner. For fair comparison, all PSNR and SSIM scores reported in the main document are calculated on the RGB channels. The results of other methods are directly cited from the original papers or generated using the official models. For the results on datasets that the authors did not report or test, we retrain their models using the official code provided by the authors.

\section{More Experiment Results}
We present more experiment results on unsupervised deraining performance and generalization performance to further elucidate the effectiveness of the proposed CSUD.
\subsection{Unsupervised Deraining Results}
% \section{Qualitative Results}
% \setcounter{figure}{0}
% \renewcommand{\thefigure}{B\arabic{figure}}
We provide additional visual comparisons on benchmark datasets in \cref{fig.SM_rain100l_results}, \cref{fig.SM_Rain12}, and \cref{fig.SM_Rain800}. We compare our CSUD with several recent state-of-the-art unsupervised and supervised image deraining methods, including \cite{MPRNet, PromptIR, DerainCycleGAN, DCDGAN, NLCL}. As shown in the figures, it can be seen that our CSUD achieves better results in removing rain streaks compared to other unsupervised methods and our CSUD preserves more texture details of image background. It is worth noting that there is a certain background color offset between the input and GT images of Rain800 dataset \cite{Rain800}, however, our CSUD aims to preserve more color and texture details of image background while removing rain streaks, so our quantitative results in the main document which are are not the best.

\subsection{Generalization Deraining Results}
\begin{figure*}[t]
  \centering
  \setlength{\abovecaptionskip}{0.1cm}
  \includegraphics[width=1\textwidth]{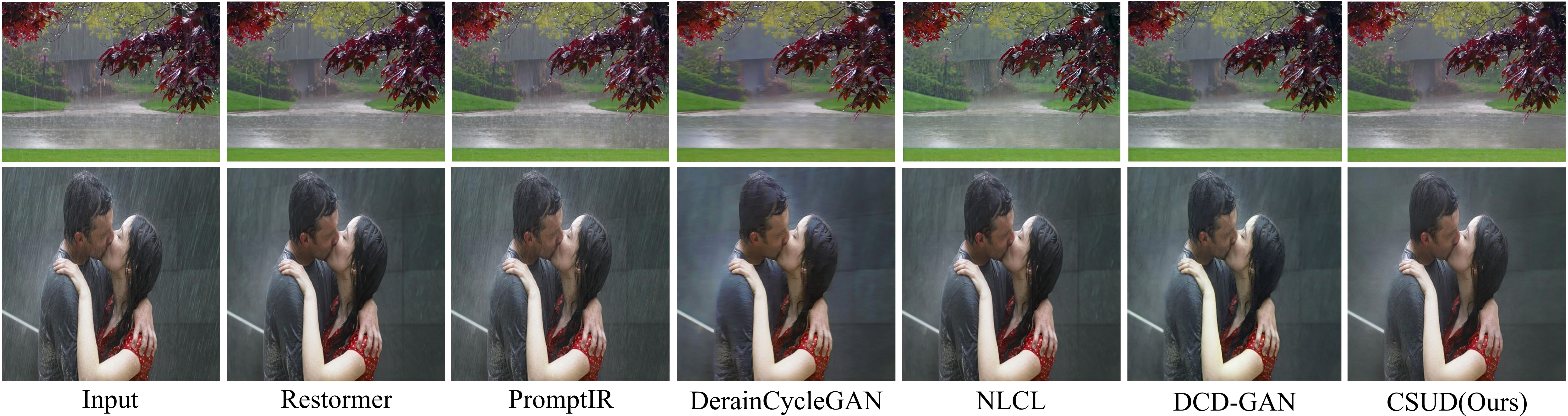}
  %\fbox{\rule[-.5cm]{0cm}{4cm} \rule[-.5cm]{4cm}{0cm}}
  \caption{Qualitative generalization results on Internet-Data \cite{SIRR} dataset.}
  \label{fig.SM_Internet}
\end{figure*}

\begin{figure*}[t]
  \centering
  \setlength{\abovecaptionskip}{0.1cm}
  \includegraphics[width=1\textwidth]{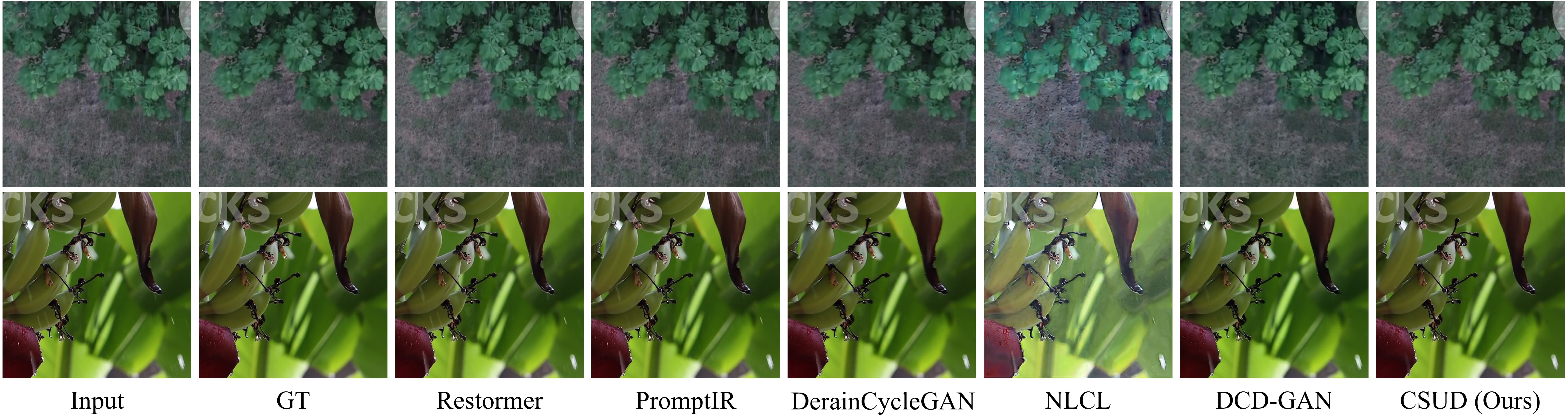}
  %\fbox{\rule[-.5cm]{0cm}{4cm} \rule[-.5cm]{4cm}{0cm}}
  \caption{Qualitative generalization results on SPA-data \cite{SPANet} dataset.}
  \label{fig.SM_SPA}
\end{figure*}

\begin{figure*}[t]
  \centering
  \setlength{\abovecaptionskip}{0.1cm}
  \includegraphics[width=1\textwidth]{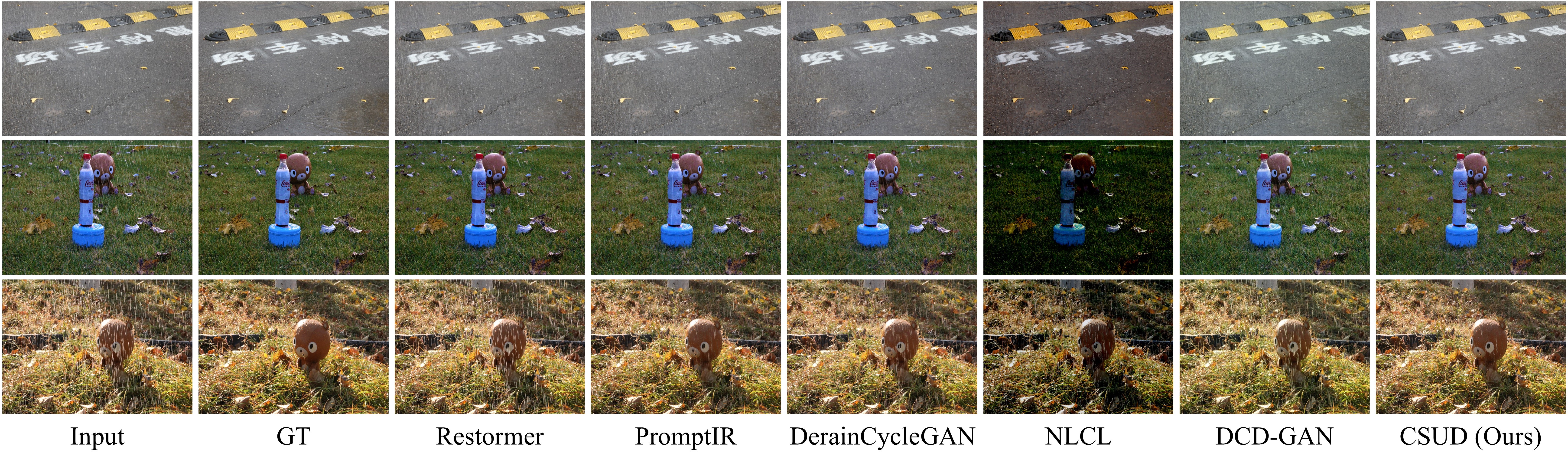}
  %\fbox{\rule[-.5cm]{0cm}{4cm} \rule[-.5cm]{4cm}{0cm}}
  \caption{Qualitative generalization results on RainDS \cite{RainDS} dataset.}
  \label{fig.SM_RainDS}
\end{figure*}

To validate the generalization capability of CSUD, we provide more additional visual comparisons with other unsupervised and supervised deraining methods, including \cite{restormer, PromptIR, DerainCycleGAN, DCDGAN, NLCL} in \cref{fig.SM_RainDS}, \cref{fig.SM_SPA}, and \cref{fig.SM_Internet}. All methods are trained on synthetic datasets and tested on the unseen real-world datasets. Compared to other methods, our CSUD achieves better visual results in real-world scenarios, which demonstrates the excellent generalization capability of CSUD.

\begin{table}[h]
\scriptsize
  \caption{Ablation experiments on the numbers of GANs. All models in the table are trained on Rain100L. SPA-Data and RealRain1K-L are used to evaluate the model’s
generalization capability. Bold fonts indicate the highest metrics. }
   \resizebox{0.48\textwidth}{!}{
   \label{tab.Numbers of GANs}
  \centering
  \begin{tabular}{c|ccc}
    \toprule
   \multirow{2}{*}{ Num of GANs }     & Rain100L         &RealRain1K-L   & SPA-data \\
    % \midrule
     & PSNR $\uparrow$ / SSIM $\uparrow$& PSNR $\uparrow$ / SSIM $\uparrow$& PSNR $\uparrow$ / SSIM $\uparrow$\\
    \midrule
   1  & 31.87 / 0.919 & 28.11 / 0.906  &33.13 / 0.932 \\
		2  & 32.92 / 0.948 & 29.08 / 0.923   &\textbf{33.67} / 0.936 \\
        %  \rowcolor{blue!8}
		4     & \textbf{33.28 / 0.954} & \textbf{29.21 / 0.928}  &33.57 / \textbf{0.939} \\
    \bottomrule
  \end{tabular}
  }
\end{table}

\subsection{More Ablation Studies}
\textbf{Effect of CSUD framework on perceptual quality.} To more comprehensively evaluate the performance of our CSUD in real world, we select 3 deraining baselines MPRNet \cite{MPRNet}, NeRD-Rain-S \cite{NeRD}, and NAFNet \cite{NAFNet}, and we use additional perceptual quality metrics to test their supervised version and unsupervised version with our CSUD on 3 real-world datasets. The perceptual quality metrics includes full-reference metrics: LPIPS~\cite{LPIPS}, DISTS~\cite{DISTS} and no-reference metric: NIQE \cite{NIQE}. As shown in \cref{table.generalization_perceptual}, our CSUD achieves the best LPIPS and DISTS with all the 3 baselines, and NeRD-Rain-S with CSUD maintains best results for all the 3 perceptual metrics on all the 3 datasets. This shows that the derained image obtained by our method can obtain higher perceptual quality, and further demonstrates the effectiveness and the generalization ability of our methods.

\textbf{Effect of the additional 3 GAN constraints.} The introduction of the additional 3 adversarial constraints aims to enhance the training stability and improve the network’s performance. To validate the necessity, we respectively train the model with 1, 2, and 4 adversarial constraints, with results shown in \cref{tab.Numbers of GANs}. It is obvious that when 4 GAN constraints are used, the deraining performance and generalization ability of the network are the best, demonstrating the effectiveness of the additional 3 GAN constraints. Note that, during inference, only the derainer is used and our framework does not introduce any additional inference overhead.

\begin{table}[h]
\scriptsize
  \caption{Ablation experiments on separation training of CSUD framework. All models in the table are trained on Rain100L. Bold fonts indicate the highest metrics. }
   \resizebox{0.48\textwidth}{!}{
   \label{tab.separation training}
  \centering
  \begin{tabular}{c|ccc}
    \toprule
   \multirow{2}{*}{ Training Strategy }     & Rain100L         &RealRain1K-L   & SPA-data \\
    % \midrule
     & PSNR $\uparrow$ / SSIM $\uparrow$& PSNR $\uparrow$ / SSIM $\uparrow$& PSNR $\uparrow$ / SSIM $\uparrow$\\
    \midrule
         Separate Training & 31.06 / 0.947 & 29.06 / \textbf{0.928}  & 32.39 / 0.936\\
        %  \rowcolor{blue!8}
		Joint Training & \textbf{33.28 / 0.954} & \textbf{29.21 / 0.928}     &\textbf{33.57 / 0.939} \\
    \bottomrule
  \end{tabular}
  }
\end{table}

\textbf{Separation training of our framework.} In order to further explore whether our unsupervised framework can train the generator and the derainer separately, we first train the generator separately and then train the derainer with the pseudo-paired rain-clean image generated by the trained generator. As shown in \cref{tab.separation training}, although it can still achieve good performance under separate training, its deraining performance and generalization ability have significantly decreased compared to joint training. Many components in our framework rely on the collaborative interaction to make derainer and generator mutually enhance each other. If the generator and derainer are trained separately, the SRloss for derainer and additional GAN constraints cannot be added to training process, and the generator cannot continuously generate pseudo-paired data, which will cause reduced constraints and performance degradation.

\section{Discussion and Limitations}
Like other image deraining methods, our method also face the same problem that it may mistakenly remove some background textures similar to the rain streaks in real rainy images, this shortcoming needs to be further improved. Additionally, our method can be widely applied in many applications such as autonomous vehicles and video surveillance. Therefore, one should be cautious of questionable results and avoid infringement of privacy or negative impact on society.

\end{document}